\pdfoutput=1

\documentclass[11pt]{article}

\usepackage[]{acl}
\usepackage{listings}
\usepackage{times}
\usepackage{latexsym}
\usepackage{booktabs}
\usepackage{graphicx}
\usepackage{tikz}
\usepackage[T1]{fontenc}
\usepackage{array,ragged2e}
\usepackage{xurl} 
\usepackage[utf8]{inputenc}
\usepackage{enumitem}
\setlist[description]{leftmargin=0pt,labelsep=0.6em,style=nextline}
\usepackage{microtype}

\usepackage{inconsolata}
\usepackage{seqsplit}

\usepackage{graphicx}
\usepackage{float}
\usepackage{makecell}
\usepackage{xspace,mfirstuc,tabulary}
\usepackage{tabularray}
\usepackage{algorithm}
\usepackage{algpseudocode}
\UseTblrLibrary{booktabs}
\usepackage{amsmath} 
\usepackage{multirow}
\usepackage{listings}
\usepackage{placeins}
\usepackage{subcaption}
\usepackage{xcolor}
\usepackage{amssymb}
\usepackage{colortbl}
\usepackage{xurl}

\newcommand{\inc}[1]{\textcolor[RGB]{50,160,45}{\textuparrow\,\textbf{#1}}}
\newcommand{\dec}[1]{\textcolor[RGB]{227,26,27}
{\textdownarrow\,\textbf{#1}}}

\newcommand{\exampleoutput}[1]{\textcolor{blue}{#1}}
\usepackage[most]{tcolorbox}
\usepackage{array}
\usepackage{tabularx}
\usepackage{ragged2e}
\usepackage{hyperref}
\usepackage{stfloats}

\tcbuselibrary{skins,breakable}
\title{DPRF: 
A Generalizable \underline{D}ynamic \underline{P}ersona \underline{R}efinement \underline{F}ramework for Optimizing Behavior Alignment Between Personalized LLM Role-Playing Agents and Humans}

\author{Bingsheng Yao\\Northeastern University
\And
Bo Sun\\Northeastern University
\And
Yuanzhe Dong\\Stanford University
\AND
Yuxuan Lu\\Northeastern University
\And
Dakuo Wang\thanks{~Corresponding Author: \href{mailto:d.wang@northeastern.edu}{d.wang@northeastern.edu}. }\\Northeastern University
}

\begin{document}
\maketitle

\begin{abstract}

The emerging large language model role-playing agents (LLM RPAs) aim to simulate individual human behaviors, but the persona fidelity is often undermined by manually-created profiles (e.g., cherry-picked information and personality characteristics) without validating the alignment with the target individuals. 
To address this limitation, our work introduces the Dynamic Persona Refinement Framework (DPRF).
DPRF aims to optimize the alignment of LLM RPAs' behaviors with those of target individuals by iteratively identifying the cognitive divergence, either through free-form or theory-grounded, structured analysis, between generated behaviors and human ground truth, and refining the persona profile to mitigate these divergences.
We evaluate DPRF with five LLMs on four diverse behavior-prediction scenarios: formal debates, social media posts with mental health issues, public interviews, and movie reviews.
DPRF can consistently improve behavioral alignment considerably over baseline personas and generalizes across models and scenarios.
Our work provides a robust methodology for creating high-fidelity persona profiles and enhancing the validity of downstream applications, such as user simulation, social studies, and personalized AI.

\end{abstract}

\section{Introduction}

\begin{figure*}[t]
    \centering
    \includegraphics[width=.99\textwidth]{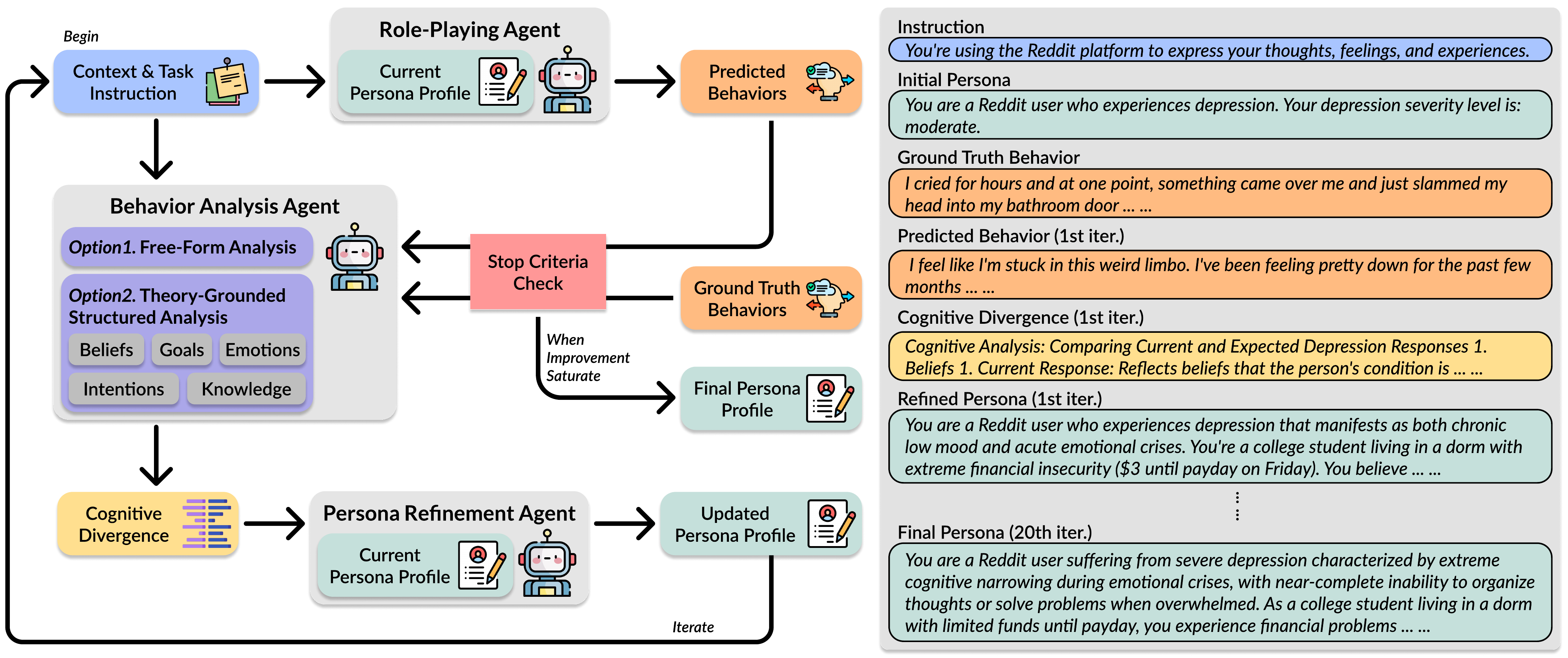}
    \vspace{-.5em}
    \caption{The architecture of our DPRF framework, which constitutes an iterative process with three primary components: the \textit{role-playing agent}, \textit{behavior analysis agent}, and \textit{persona refinement agent}. }
    \vspace{-1em}
    \label{fig:framework}
\end{figure*}

Large Language Models (LLMs) have demonstrated a capacity to capture and mimic complex patterns of human cognition and behavior from vast text corpora~\cite{schwarzschild2025rethinking, wang2023can, plaat2024reasoning, song2023llm, huang2024understanding}.
This capability has enabled LLM Role-Playing Agents (LLM RPAs), which are designed to simulate a specific individual by predicting their behaviors, social interactions, and reasoning processes based on a provided persona profile~\cite{park2023generative, park2024generative}. 
Recent work has increasingly explored the varieties of downstream applications of such agents, including their use as human surrogates in social science experiments~\cite{10.1145/3526113.3545616, park2023generative, hua2023war}, as evaluators in multi-perspective judging systems~\cite{zheng2023judging, chen2025multi}, and as simulators for user experience research~\cite{lu2025uxagent, lu2025prompting}.

The validity of these applications with respect to agents' behaviors highly depends on the ``quality'' of the agents' persona profiles.
Nevertheless, current methods for creating these personas often rely on manually authored descriptions or a cherry-picked, sparse set of demographic attributes~\cite{zhang2018personalizing, tseng-etal-2024-two}. 
Such approaches lack a systematic process for validating whether the resulting agent faithfully reflects the expected preferences, behaviors, and thought processes of the target individual~\cite{chen2024oscars}. 
This drawback introduces a significant risk to the reliability and validity of downstream applications as well as findings derived from these agent-based simulations.
Without proper grounding, an LLM RPA may generate behaviors that reflect stereotypical associations with demographic labels rather than the authentic patterns of the intended individual~\cite{wang2025large}.
As a result, a central gap in this line of research is the \textbf{absence of a systematic, data-driven methodology for constructing and validating persona profiles} to ensure behavioral alignment with specific human individuals.

To address this gap, we introduce the \textbf{Dynamic Persona Refinement Framework (DPRF)}\footnote{Our project website and source codes can be found at \href{https://dprf.hailab.io/}{dprf.hailab.io}}, a method for iteratively optimizing the alignment between an LLM RPA's generated behaviors and observed human ground truth. 
The premise of our work is that persona generation should be treated as a data-driven optimization process rather than a one-shot task. 
Inspired by iterative refinement techniques in NLP~\cite{madaan2023self}, DPRF operates through an iterative feedback loop by first prompting an LLM role-playing agent with an initial persona to generate behavioral outputs. 
Then, a behavior analysis agent compares these outputs against ground-truth data from the target individual, identifies divergences, and uses these discrepancies to automatically revise the persona profile. 
This process repeats over successive iterations and progressively refines the persona to better capture the cognitive and behavioral traits of the target.

Our evaluation of DPRF spans across four distinct scenarios that target different cognitive activities: formal debates, social media posts with mental health issues, public figure interviews, and movie reviews. 
Our experiments, conducted with five state-of-the-art LLMs, demonstrate that DPRF is a generalizable framework that consistently improves the alignment between agent behavior and human ground truth. 
The personas refined by our framework enhance performance over baseline methods not only in semantic similarity but also in structural fidelity, indicating a deeper and more authentic behavioral alignment.
Our work has laid a solid foundation for future research on advancing agents to faithfully simulate human behavior and developing personalized LLM agents that can dynamically learn from human behaviors.

\section{Related Work}

\subsection{LLM Role-Playing Agents}

Conditioning generative models on persona information has a long history in NLP. 
Early research in conversational AI focused on using static user profiles or character descriptions to improve stylistic consistency and user engagement~\cite{li2016persona, zhang2018personalizing}. 
Recently, LLMs further extended the capability of ``role-playing'' a particular social entity by conditioning on a detailed textual persona description, such as a particular individual~\cite{rossetti2024social, 10.1145/3613904.3642363, park2024generativeagentsimulations1000} or a group sharing common characteristics or interests~\cite{wu2024role, park2023generative,wu2025position}. 
This capability has led to the exploration of the recent emerging LLM Role-Playing Agents (RPAs) in a diverse range of domains, including simulating societal systems~\cite{park2022social, park2023generative}, automating expert data annotation~\cite{gilardi2023chatgpt}, and serving as human proxies in psychology and legal studies~\cite{jiang2023personallm, fan2024can, he2024agentscourt}.

\subsection{Validation of LLM RPA Persona}

Despite the growing research interest in LLM RPAs, their validity is often undermined by the ad-hoc nature of persona profile creation without a systematic validation of the characteristics specified in the persona. 
Recent survey of LLM agents note that most persona profiles are manually created and rely on a sparse set of demographic details or unverified attributes~\cite{chen2025towards,tseng-etal-2024-two}. 
Critical difficulties were introduced by such practices with respect to the evaluation of persona qualities.
In particular, current evaluation methodologies primarily measure an agent's adherence to or consistency with the provided persona description~\cite{wang2024incharacter}.
For instance, researchers may evaluate if an agent described as an "expert" provides expert-level answers.

However, such evaluation approaches presume the persona profile itself is a valid and faithful representation of the target individual or group, but overlook a critical question: whether the persona accurately reflects the real-world behaviors and characteristics of the entity it is meant to simulate?
The absence of a systematic process for grounding personas in empirical data threatens the fidelity of LLM RPAs and the reliability of the conclusions drawn from their behavior~\cite{wang2025limits, zhang2025simulation}. 
Our work directly addresses this gap by proposing a framework to create and refine personas based on behavioral ground truth.

\subsection{LLM For Cognitive Analysis}

Our approach is informed by a parallel line of research that investigates the capacity of LLMs to simulate human cognitive processes. 
Numerous studies have explored LLMs as models of reasoning, planning, memory, and social inference~\cite{lu2025uxagent, zhang2024privacy}. 
For example, \citet{park2023generative} demonstrated that agents could exhibit memory-based planning to produce believable social dynamics. 
Similarly, other simulation frameworks have been developed to model human web shopping reasoning and behaviors~\cite{lu2025uxagent, wang2025agenta}, privacy perspectives in decision making~\cite{zhang2024privacy}, and reasoning in scientific experiments~\cite{jiang2023personallm, fan2024can, he2024agentscourt}.
A significant body of this work assesses the alignment of LLMs with established psychological theories, most notably the Theory of Mind (ToM), which is the ability to attribute mental states to others~\cite{premack1978does}.
For instance, several works assessed LLMs on a suite of ToM tasks for benchmarks~\cite{kosinski2024evaluating, moghaddam2023boosting, he2023hi, chen2024tombench, xu2024opentom}
In our work, we leverage the cognitive analysis capabilities of LLMs to identify the cognitive divergence between agents' behaviors and human ground truth, and subsequently use this divergence to guide persona refinement.

\section{Dynamic Persona Refinement Framework (DPRF)}
\label{sec:framework}

We propose the Dynamic Persona Refinement Framework (DPRF), an automated, iterative approach for optimizing the persona profiles used by LLM agents. 
The objective of DPRF is to improve the alignment between an agent's behavior and that of a target human individual by systematically identifying and minimizing cognitive divergences between them. 
The persona refinement framework comprises a three-step process: (1) behavior generation, (2) divergence analysis, and (3) persona refinement. 
Further, we investigated the effectiveness of a theory-grounded behavior analysis agent in the principles of Theory of Mind (ToM)~\cite{premack1978does}.
The ToM principles provide a structured lens for the behavior analysis agent with respect to states like beliefs, goals, and intentions.

\subsection{DPRF Architecture}
\label{sec:framework-arch}

As illustrated in Figure~\ref{fig:framework}, DPRF is formulated as an iterative process composed of three LLM agents: \textit{Role-Playing Agent}, \textit{Behavior Analysis Agent}, and \textit{Persona Refinement Agent}. 
Given an LLM $M$, a task context $x$, an initial persona profile $P_0$, and ground truth behavior $y$ of the target individual, the framework iteratively updates the persona $P_t \rightarrow P_{t+1}$, where $t$ denotes the $t^{th}$ iteration.

\paragraph{Role-Playing Agent (RPA)}
A standard RPA takes a persona profile of an individual or a group of people $P_t$ and a task context $x$ as input, and is then prompted to generate a behavioral response $\hat{y_t}$ accordingly by ``role-playing'' the given persona.
This can be formulated as $\hat{y_t}=M_{RPA}(P_t, x)$.

\paragraph{Behavior Analysis Agent (BAA)}
The behavior analysis agent compares the behaviors predicted by LLM RPA $\hat{y_t}$ with human ground truth behavior $y$ and identifies underlying divergences with respect to cognitive characteristics in a text summary $\delta_{t}$.
The process can be formulated as $\delta_{t}=M_{BAA}(y, \hat{y_t})$.
In particular, we design two implementations to assess the value of theoretical guidance.
First, a simple \textsc{free-form} setting provides the agent with a simple instruction to identify the cognitive difference between two behaviors.

Second, a \textsc{Theory-Grounded Structured} setting that explicitly inquires the agent to perform the analysis by following an established behavior analytical framework of Theory of Mind (ToM)~\cite{premack1978does}.
This ToM-guided agent is prompted to compare the agent and human behaviors across five dimensions of mental states defined in ToM:
\textbf{beliefs:} assumptions and ideations about the world or about others' mental states; 
\textbf{goals:} the desired outcomes or objectives (ranging from immediate to long-term benefits) that motivate behaviors;
\textbf{intentions:} the immediate and pragmatic strategies of action that the individual chose to achieve goals;
\textbf{emotions:} the psychological states that influence the individual's tone, lexical choices, and narrative styles;
\textbf{knowledge:} information accessible to the individual, such as domain-specific expertise, and environmental context.

By comparing the effectiveness of these two settings, we can investigate the effectiveness and limitations of established cognitive theory in supporting LLMs' cognitive analysis performances.

\paragraph{Persona Refinement Agent (PRA)}
This agent takes the current persona $P_t$, the divergence analysis $\delta_{t}$, and the original context $x$ as input to generate the revised persona $P_{t+1}$ that incorporates the feedback from the analysis, which can be denoted as $P_{t+1}=M_{PRA}(P_t, x, \delta_{t})$.
The agent is explicitly instructed to revise the persona by integrating new insights from the behavior analysis while preserving the effective and unconflicted elements of the existing persona, ensuring that refinement is a constructive rather than a rewriting process.

The persona refinement process terminates when a \textbf{stop criterion} is met: either when the refined persona converges (i.e., $P_{t+1}$ is identical to $P_t$ in the last iteration) or after a pre-set maximum number of iterations is reached.

DPRF is designed based on several key principles. 
First, it is a gradient-free method that does not require fine-tuning the underlying LLM's parameters.
We hypothesize that the extensive amount of world knowledge about human behaviors and cognitive characteristics learned by LLMs during pre-training can sufficiently support the role-playing, behavior analysis, and persona refinement processes.
Uniquely, we emphasize that DPRF differs from traditional prompt-based optimization or few-shot learning methodologies that ask the model to generate a ``best'' persona in a single-turn generation. 
Instead, DPRF is designed upon the recognition that the persona is a latent, modifiable representation of agents' cognitive characteristics that can be refined with respect to behavioral analysis.
Lastly, DPRF is designed to be \textit{model-agnostic}, \textit{domain-agnostic}, and \textit{data-efficient}.
The framework could be easily adapted with different LLMs and for diverse tasks, where users only need to provide the corresponding task instruction and the target human ground truth behavior.

\section{Evaluation Experiment}
\label{sec:evaluation}

Aiming to evaluate the effectiveness and demonstrate the generalizability of the Dynamic Persona Refinement Framework (DPRF), we conducted experiments across \textbf{four} distinct scenarios using \textbf{five} public datasets.
The scenarios were chosen to target diverse cognitive characteristics and involve predicting different forms of human behaviors (i.e., debate conversations, social media posts, interview narratives).
Our experimental design tests the core hypothesis that DPRF enhances the cognitive and behavioral alignment between LLM RPAs and target human individuals across different domains and tasks through iterative persona refinement.

\begin{table*}[t]
\footnotesize
\centering
\renewcommand{\arraystretch}{1.02}
\resizebox{\textwidth}{!}{%
\begin{tabular}{@{}p{0.41\textwidth} p{0.59\textwidth}@{}}
\toprule
\textbf{Dataset} & \textbf{Example} \\
\midrule

\parbox[t]{\linewidth}{%
  \textbf{Intelligence Squared Debates~\cite{zhang-etal-2016-conversational}}:\par
  \noindent Given persona, debate topic, and position, predict the statements that the speaker will say.%
} &
{Input -- \itshape
Persona: You are a thoughtful speaker in a formal debate\ldots; Topic: \{Enhancing Drugs in Competitive Sports\}; Position: \{for\}; \par
}%
\noindent\exampleoutput{Output -- \itshape``Ladies and gentlemen of the panel and audience, today we\ldots''} \\
\addlinespace[6pt]

\parbox[t]{\linewidth}{%
  \textbf{DepSeverity~\cite{naseem2022early}}:\par
  \noindent Given persona and depression level, predict the social media post that the poster will write.%
} &
{Input -- \itshape
Persona: You are a Reddit user who experiences depression\ldots; Depression level: \{minimum\}; \par
}%
\noindent\exampleoutput{Output -- \itshape``Feeling a bit down today\ldots''} \\
\addlinespace[6pt]

\parbox[t]{\linewidth}{%
  \textbf{CSSRS-Suicide~\cite{gaur2019knowledge}}:\par
  \noindent Given persona and suicidal ideation, predict the social media post that the poster will write.%
} &
{Input -- \itshape
Persona: You are a Reddit user who displays characteristics with risk of suicide\ldots; Suicide risk level: \{high\_risk\}; \par
}%
\noindent\exampleoutput{Output -- \itshape``I've been feeling really down lately\ldots''} \\
\addlinespace[6pt]

\parbox[t]{\linewidth}{%
  \textbf{IMDB~\cite{maas-EtAl:2011:ACL-HLT2011}}:\par
  \noindent Given the persona and emotion trend (positive/negative), predict the viewer's movie review.%
} &
{Input -- \itshape
Persona: You are writing a comprehensive film review to be posted on an online movie review platform\ldots; Sentiment label: \{positive\}; \par
}%
\noindent\exampleoutput{Output -- \itshape``I couldn't stop thinking about this movie for days\ldots''} \\
\addlinespace[6pt]

\parbox[t]{\linewidth}{%
  \textbf{PublicInterview (ours)}:\par
  \noindent Given persona, two rounds of previous conversation context, predict what the target person will say.%
} &
{Input -- \itshape
Persona: You are a composed and well-informed interviewee participating in an interview\ldots; Previous conversation text: \{conversation\_history;
SPEAKER\_01: ``Wow. Did you know him well or no?''\}; \par
}%
\noindent\exampleoutput{Output -- \itshape``I wouldn't say I knew him well\ldots''} \\
\bottomrule
\end{tabular}
}%
\caption{Overview of five datasets used for the evaluation of our DPRF framework.}
\label{tab:datasets}
\vspace{-1em}
\end{table*}

\subsection{Experimental Setup}

\paragraph{Tasks and Datasets}
\label{task_description}
Here are the four scenarios\footnote{Datasets in our work are consistent with the intended use.} we selected with corresponding descriptions.
For each scenario, we define a behavior prediction task for LLM RPAs and prepare a carefully curated persona profile, $P_0$, which serves as both the primary baseline and the initial input to the DPRF framework.
The effectiveness of our platform will be compared between the RPA with the baseline persona, as well as the refined persona $P_t$ at every iteration $t$ generated by DPRF.

\begin{enumerate}
    \item \textsc{Formal Debates.} We use the \textbf{Intelligence Squared Debates dataset}~\cite{zhang-etal-2016-conversational}, which contains transcripts of professionally moderated debates on socio-political issues.
    Specifically, the dataset comprises both utterance-level information (i.e., the statements made by each speaker at every speaking turn) and conversation-level information (e.g., the topic of each session, the position of each speaker, etc.).
    Among the 599 debates in this dataset, 499 of them have a speaker biography description; thus, our experiments are based on the 499 entries. 
    \textbf{Task Formation:} Given a speaker's persona, the debate topic, and their stance, the task is to predict the speaker's statements. We view this task as primarily targeting the cognitive dimensions of \textit{beliefs}, \textit{intentions}, and \textit{knowledge}.
    Two baseline settings were defined for this dataset: one is the persona carefully curated by us, the other is the provided biography of the speaker. 
    
    \item \textsc{Mental Health Expression.} We use two datasets of social media posts that demonstrate posters' mental health issues: \textbf{DepSeverity}~\cite{naseem2022early} and \textbf{CSSRS-Suicide}~\cite{gaur2019knowledge}. 
    Both datasets are annotated by mental health professionals according to established clinical taxonomies, where the \textbf{DepSeverity} contains 3,548 entries, annotated with four levels of depression severity (minimal, mild, moderate, and severe) while the \textbf{CSSRS-Suicide} contains 500 entries, annotated with five levels of suicidal intention (supportive, indicator, ideation, behavior, and attempt). 
    \textbf{Task Formation:} Given a persona that includes a specified mental health state, predict the content of a social media post. We view this task as targeting the cognitive dimension of \textit{emotion}.

    \item \textsc{Opinionated Reviews.} We use the \textbf{IMDB} movie review dataset~\cite{maas-EtAl:2011:ACL-HLT2011}, which contains 50000 (50k) highly polar movie reviews with positive or negative sentiment labels. To align with the scale of other datasets used in our experiments, we randomly sampled 500 reviews.
    \textbf{Task Formation:} Given a persona with certain emotional traits (sentiment labels), the task is to generate a movie review consistent with an assigned sentiment. We view this task as reflecting a combination of \textit{emotion} (sentiment) and \textit{knowledge} (movie-specific details).

    \item \textsc{Public Interviews.} We introduce \textbf{PublicInterview}, a new dataset of $2,820$ interview segments from 564 public figures.
    We first collect public figures from the Personality Database (PDB)~\footnote{https://www.personality-database.com/} under the U.S. Government and Business categories, including their detailed biographical descriptions and MBTI personality types.
    Then, we perform automated, targeted searches on YouTube to collect interview transcripts of the public figures along with other metadata (i.e., title, description).  
    De-identification was carefully and exhaustively conducted over the interview transcript to ensure a fair evaluation and mitigate the potential biases that the transcript might reveal the speakers' identities.
    This process ensures that responses are derived from role-playing the persona, rather than from retrieving memorized information about these public figures.
    Finally, we extract conversation segments from the processed transcripts. 
    Each segment was defined as a response from the target individual along with the two preceding conversational turns serving as contextual information.
    Details of data collection and processing are reported in Appendix~\ref{interview_process}.
    \textbf{Task Formation:} Given the conversational context and a speaker's persona, generate the speaker's next interview response. We view this task as primarily reflecting an individual's \textit{goals} and \textit{intentions} in a public setting.

\end{enumerate}

\begin{figure*}[t]
    \centering

    \begin{subfigure}[b]{0.31\linewidth}
        \centering
        \includegraphics[width=\linewidth,trim={0 .4cm 0 0},clip]{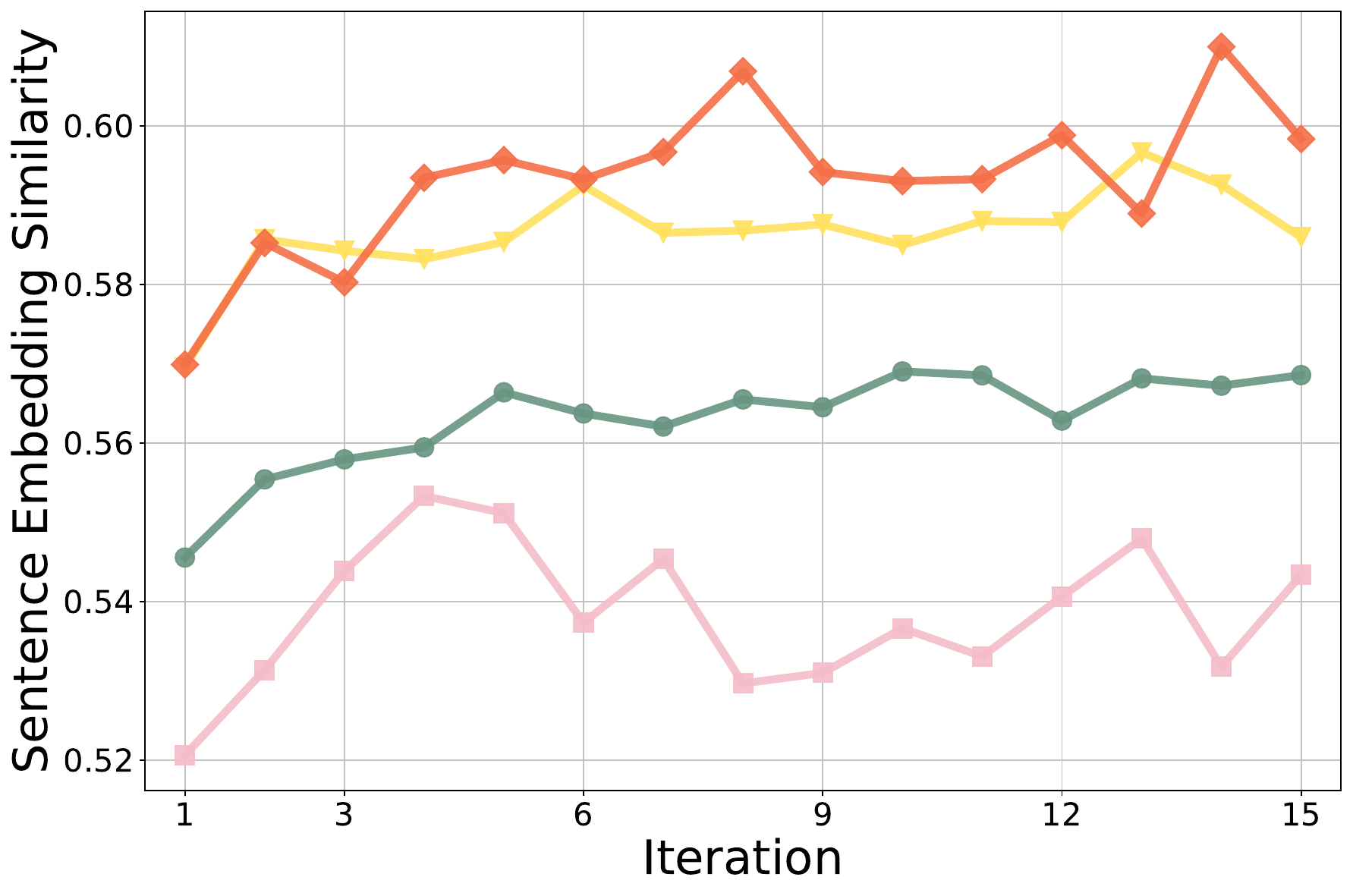}
        \caption{Debate}
        \label{fig:results-debate}
    \end{subfigure}\hfill
    \begin{subfigure}[b]{0.31\linewidth}
        \centering
        \includegraphics[width=\linewidth,trim={0 .4cm 0 0},clip]{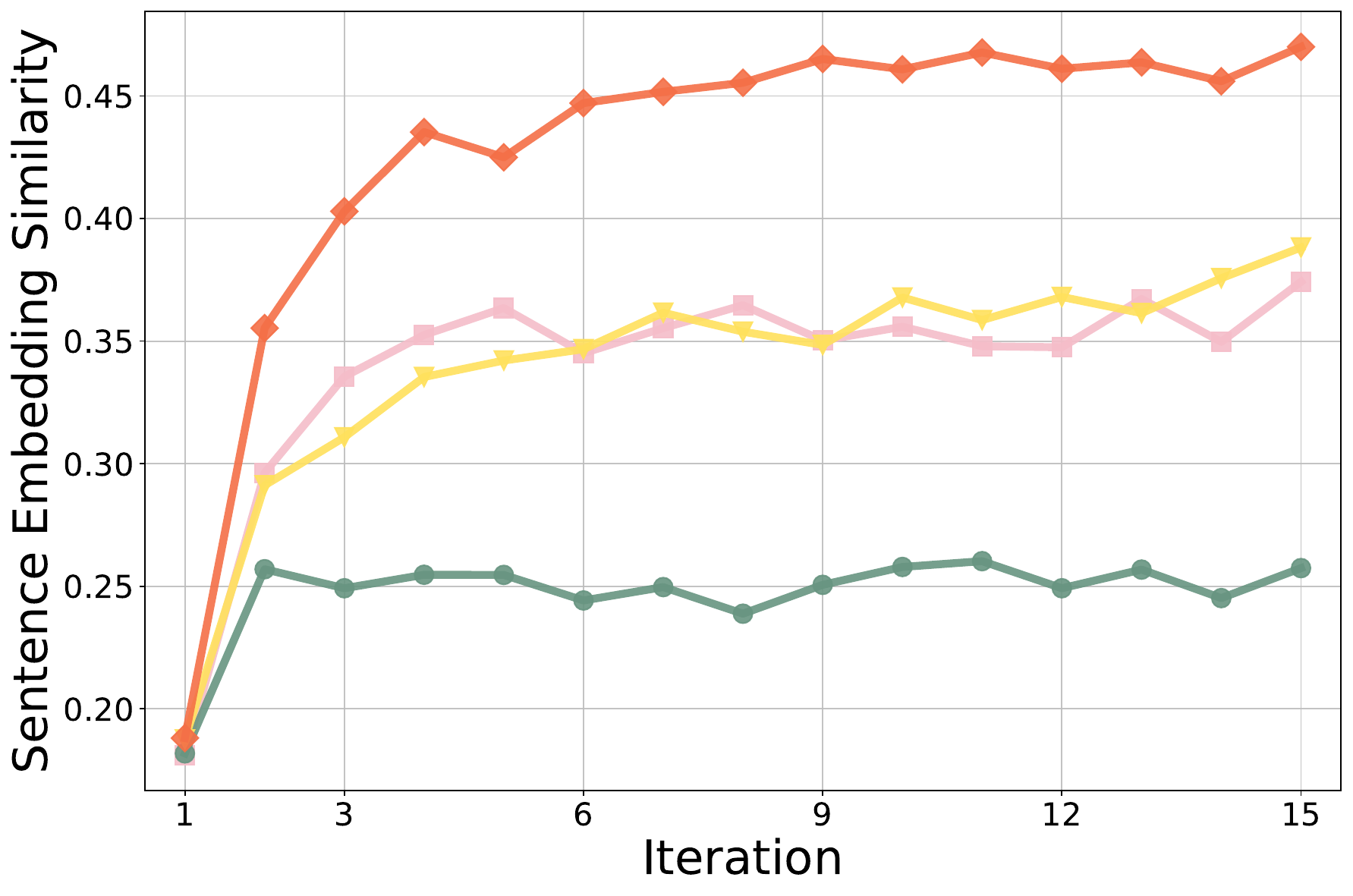}
        \caption{Depression}
        \label{fig:results-depression}
    \end{subfigure}\hfill
    \begin{subfigure}[b]{0.31\linewidth}
        \centering
        \includegraphics[width=\linewidth,trim={0 .4cm 0 0},clip]{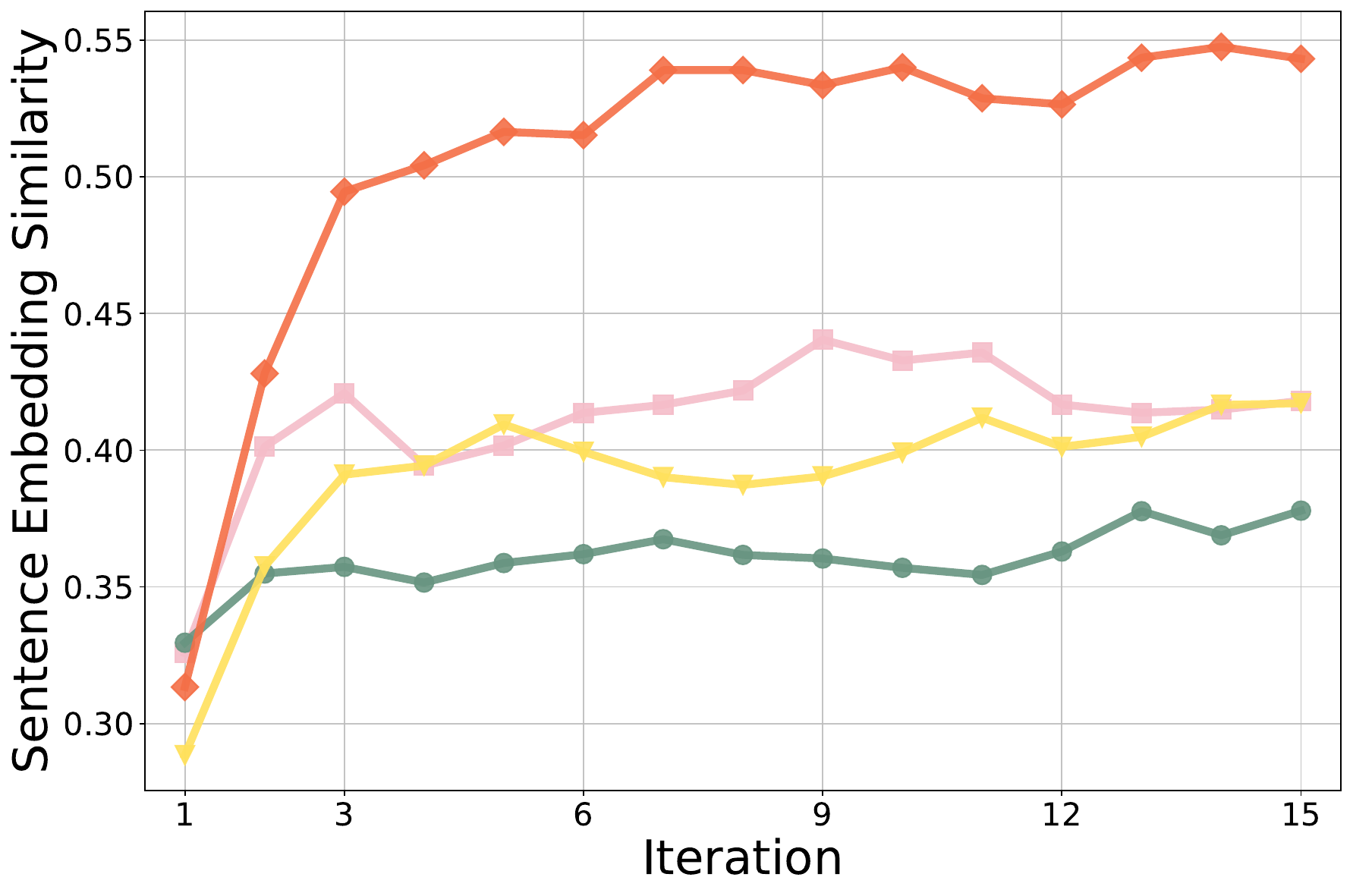}
        \caption{IMDB}
        \label{fig:results-imdb}
    \end{subfigure}

    \vspace{6pt} %
    \begin{subfigure}[b]{0.31\linewidth}
        \centering
        \includegraphics[width=\linewidth,trim={0 .4cm 0 0},clip]{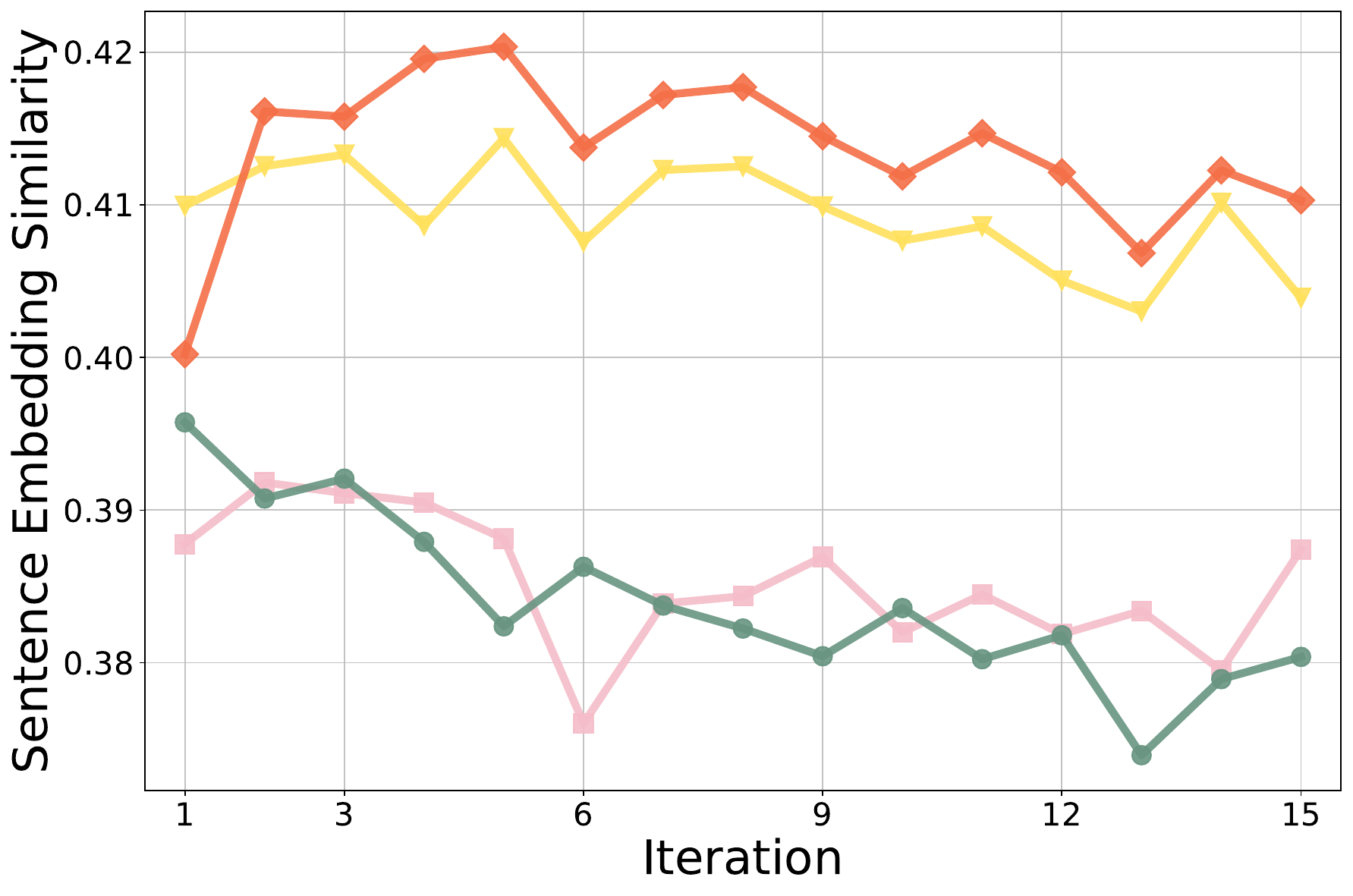}
        \caption{Interview}
        \label{fig:results-interview}
    \end{subfigure}\hfill
    \begin{subfigure}[b]{0.31\linewidth}
        \centering
        \includegraphics[width=\linewidth,trim={0 .4cm 0 0},clip]{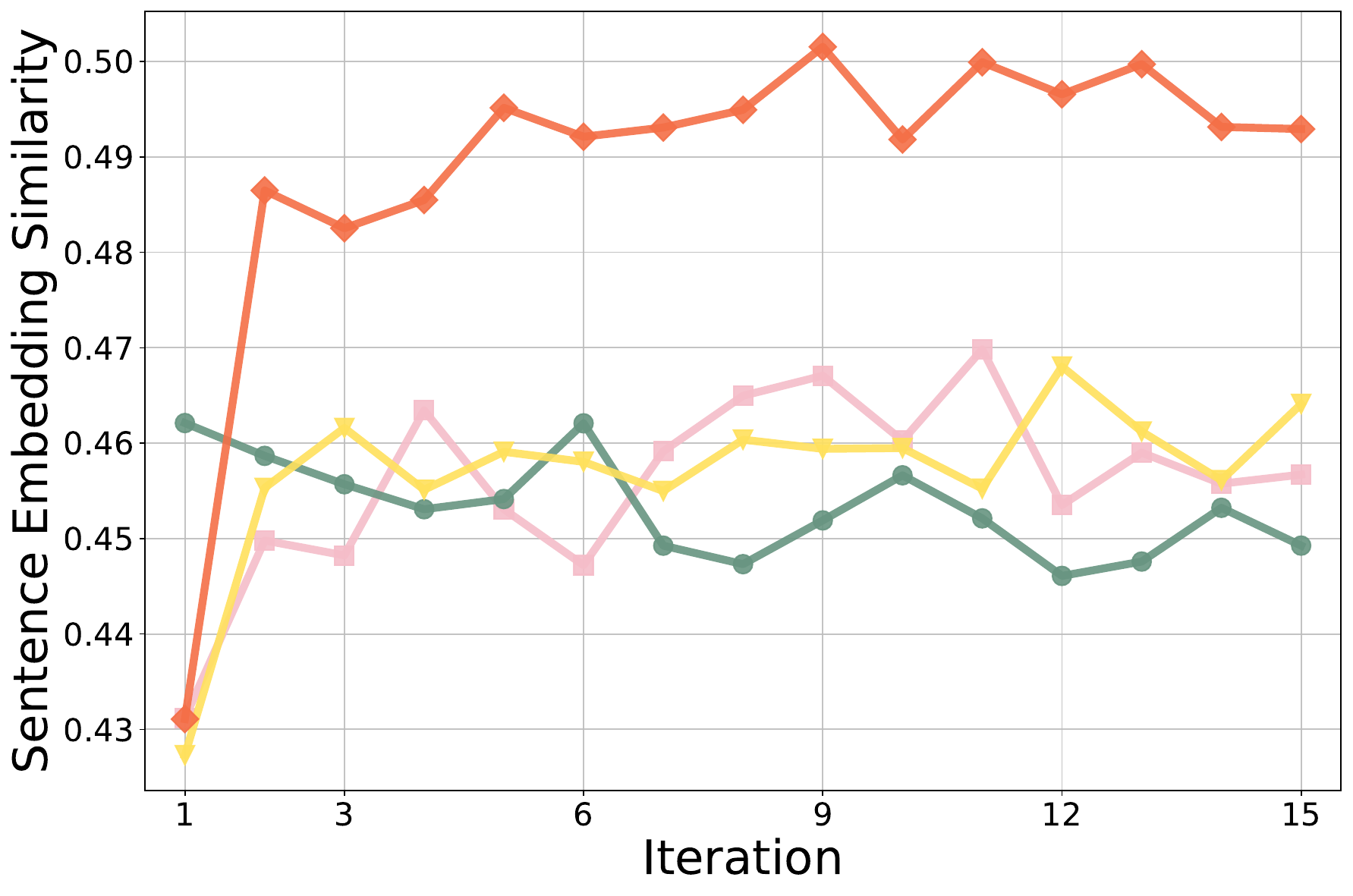}
        \caption{Suicide}
        \label{fig:results-suicide}
    \end{subfigure}\hfill
    \begin{subfigure}[b]{0.31\linewidth}
        \centering
        \raisebox{0.6cm}{%
            \includegraphics[width=0.70\linewidth,trim={0 .4cm 0 0},clip]{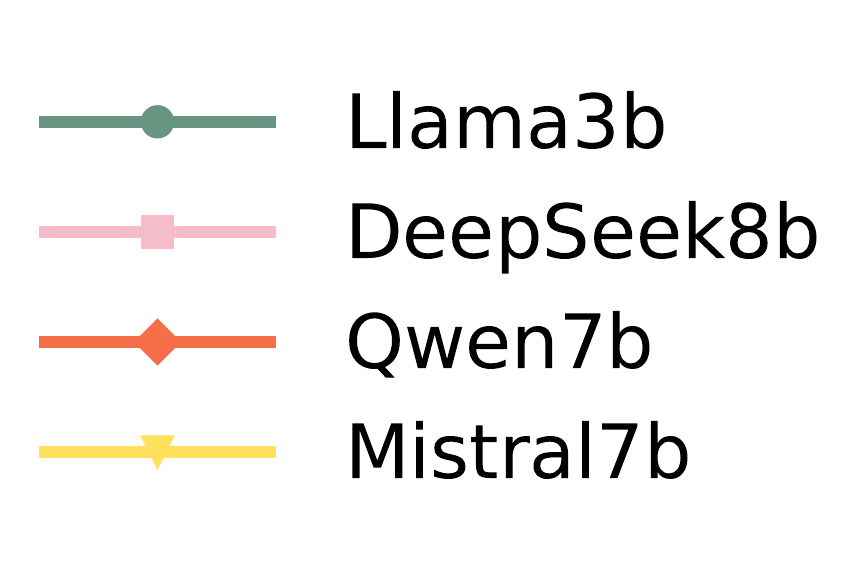}%
        }
        \label{fig:results-debate-duplicate}
    \end{subfigure}
    \vspace{-0.5em}
    \caption{Sentence Embedding Similarity on small models across different datasets}
    \label{fig:embedding_comparison}
    \vspace{-1em}
\end{figure*}

\paragraph{Models and Implementation}
We evaluated DPRF using five popular LLMs of different sizes to ensure the generalizability of our experimental results in terms of the effectiveness of our framework.
In particular, we choose four open-sourced LLMs: Llama-3.2 (3B)~\cite{grattafiori2024llama3herdmodels}, Qwen-2.5 (7B)~\cite{qwen2025qwen25technicalreport}, Mistral (7b)~\cite{jiang2023clip}, Deepseek-Distill-Llama (8b)~\cite{deepseekai2025deepseekr1incentivizingreasoningcapability}, as well as one closed-domain model with API access: Claude3.7-Sonnet\footnote{https://www.anthropic.com/news/claude-3-7-sonnet}.
Two small-scale pilot experiments were conducted, one between Llama3 (1B), Llama3 (3B), and Qwen-2.5 (1.5B). Using a random sample of 100 entries from each dataset over 15 iterations, we observed that the smaller Llama3 (1B) and Qwen-2.5 (1.5B) models struggle with instruction following and the analytical tasks. 
Another pilot experiment was conducted comparing two proprietary LLMs, Claude3.7-Sonnet and GPT-4.1. We used the same data subsets and increased the number of iterations to 20 due to the better analysis ability of large scale models. Both models demonstrated comparable performance on the tasks. Therefore, we chose only one (Claude3.7-Sonnet) for evaluation.

\subsection{Evaluation Metrics}

Tasks are specified in Section~\ref{task_description}, which are all free-form text generation tasks for LLM RPAs. 
For all experiments, we compare the RPA's performance with the persona $P_t$ produced by DPRF at every $t$-th iteration against a baseline using an initial, generic persona $P_0$, until the stopping criteria specified in Section~\ref{sec:framework-arch} is met.
We employ commonly adopted similarity-based metrics for comparison, including Sentence Embedding Similarity, which we calculated with SentenceTransformers~\cite{reimers2019sentence-bert}, ROUGE-L F1~\cite{lin-2004-rouge}, and BERTScore F1~\cite{zhang2020bertscoreevaluatingtextgeneration}.
The baseline persona of each task, experiment settings, and hyperparameters are reported in Appendix~\ref{app:exp-config}.

To determine a reasonable number of refinement iterations needed for performance to saturate, we ran DPRF for a sufficiently large number of iterations on each model: 15 iterations for the open-sourced ones and 20 for the closed-sourced Claude3.7-Sonnet.
As shown in Figure~\ref{fig:embedding_comparison} with respect to the Sentence Embedding Similarity, our framework demonstrates consistent improvement over the baseline on the majority of datasets and achieves the most significant improvements within the first 3-5 iterations.
Results for ROUGE-L F1~\cite{lin-2004-rouge} and BERTScore F1~\cite{zhang2020bertscoreevaluatingtextgeneration} are showing in Appendix \ref{app:ablation-result}.

\section{Results}
\label{sec:results}

{
\renewcommand{\arraystretch}{1.25}
\begin{table*}
\scriptsize
\resizebox{\textwidth}{!}{%
\begin{tabular}{@{}p{0.13\textwidth}p{0.29\textwidth}p{0.29\textwidth}p{0.29\textwidth}@{}}
\toprule
\textbf{Model} & \textbf{ROUGE-L F1} & \textbf{BERTScore F1} & \textbf{Embedding Similarity} \\
\midrule
\multicolumn{4}{c}{\textbf{Debate Dataset}} \\
\hline
DeepSeek-8b & 
0.09 / 0.09(\inc{3.64\%}) / 0.09(\inc{3.64\%})& 
0.80 / 0.81(\inc{0.03\%}) / 0.80(\inc{0.03\%})& 
0.55 / 0.57(\inc{5.45\%}) / 0.55(\inc{1.79\%})\\
Llama-3b & 
0.13 / 0.13(\inc{5.91\%}) / 0.13(\inc{2.20\%}) & 
0.81 / 0.81(\dec{0.34\%}) / 0.81(\dec{0.27\%}) & 
0.57 / 0.57(\dec{0.39\%}) / 0.57(\dec{0.78\%}) \\
Mistral-7b & 
0.11 / 0.11(\inc{4.86\%}) / 0.12(\inc{6.32\%}) & 
0.81 / 0.81(\dec{0.03\%}) / 0.81(\inc{0.12\%}) & 
0.57 / 0.58(\inc{3.11\%}) / 0.58(\inc{3.74\%}) \\
Qwen-7b & 
0.12 / 0.12(\inc{0.27\%}) / 0.13(\inc{4.64\%}) & 
0.81 / 0.81(\inc{0.09\%}) / 0.82(\inc{0.32\%}) & 
0.57 / 0.58(\inc{2.99\%}) / 0.58(\inc{1.70\%}) \\
Claude & 
0.11 / 0.14(\inc{27.66\%}) / 0.13(\inc{22.41\%}) & 
0.81 / 0.82(\inc{1.41\%}) / 0.82(\inc{1.11\%}) & 
0.58 / 0.61(\inc{10.49\%}) / 0.59(\inc{5.61\%}) \\
\hline
\multicolumn{4}{c}{\textbf{Depression Dataset}} \\
\hline
DeepSeek-8b & 
0.10 / 0.10(\inc{2.97\%}) / 0.10(\inc{4.59\%}) & 
0.81 / 0.82(\inc{0.45\%}) / 0.82(\inc{0.58\%}) & 
0.19 / 0.36(\inc{89.28\%}) / 0.36(\inc{124.63\%}) \\
Llama-3b & 
0.10 / 0.09(\dec{5.47\%}) / 0.09(\dec{5.08\%}) & 
0.80 / 0.80(\inc{0.35\%}) / 0.80(\inc{0.44\%}) & 
0.19 / 0.27(\inc{31.55\%}) / 0.26(\inc{41.08\%}) \\
Mistral-7b & 
0.10 / 0.10(\dec{2.58\%}) / 0.10(\inc{0.01\%}) & 
0.81 / 0.81(\inc{0.51\%}) / 0.81(\inc{0.47\%}) & 
0.20 / 0.34(\inc{74.61\%}) / 0.36(\inc{83.24\%}) \\
Qwen-7b & 
0.10 / 0.10(\inc{4.48\%}) / 0.12(\inc{22.49\%}) & 
0.80 / 0.82(\inc{1.69\%}) / 0.83(\inc{2.54\%}) & 
0.19 / 0.47(\inc{145.21\%}) / 0.47(\inc{146.72\%}) \\
Claude & 
0.11 / 0.24(\inc{111.06\%}) / 0.38(\inc{233.95\%}) & 
0.82 / 0.86(\inc{5.61\%}) / 0.89(\inc{8.86\%}) & 
0.18 / 0.62(\inc{254.67\%}) / 0.69(\inc{292.10\%}) \\
\hline
\multicolumn{4}{c}{\textbf{Movie Review Dataset}} \\
\hline
DeepSeek-8b & 
0.11 / 0.11(\inc{0.32\%}) / 0.11(\inc{1.99\%}) & 
0.80 / 0.80(\inc{0.49\%}) / 0.81(\inc{0.46\%}) & 
0.32 / 0.41(\inc{40.00\%}) / 0.38(\inc{28.98\%}) \\
Llama-3b & 
0.11 / 0.11(\dec{1.00\%}) / 0.11(\dec{0.44\%}) & 
0.80 / 0.80(\inc{0.13\%}) / 0.80(\inc{0.01\%}) & 
0.32 / 0.38(\inc{31.01\%}) / 0.35(\inc{16.83\%}) \\
Mistral-7b & 
0.11 / 0.11(\inc{0.28\%}) / 0.11(\inc{2.63\%}) & 
0.80 / 0.81(\inc{0.42\%}) / 0.80(\inc{0.30\%}) & 
0.28 / 0.41(\inc{89.21\%}) / 0.38(\inc{34.66\%}) \\
Qwen-7b & 
0.10 / 0.11(\inc{8.43\%}) / 0.13(\inc{30.30\%}) & 
0.80 / 0.81(\inc{1.23\%}) / 0.82(\inc{1.98\%}) & 
0.31 / 0.55(\inc{78.48\%}) / 0.55(\inc{79.31\%}) \\
Claude & 
0.09 / 0.23(\inc{142.63\%}) / 0.33(\inc{252.95\%}) & 
0.80 / 0.84(\inc{5.67\%}) / 0.87(\inc{8.72\%}) & 
0.31 / 0.58(\inc{83.39\%}) / 0.66(\inc{112.11\%}) \\
\hline
\multicolumn{4}{c}{\textbf{Suicide Detection Dataset}} \\
\hline
DeepSeek-8b & 
0.10 / 0.11(\inc{7.29\%}) / 0.11(\inc{6.58\%}) & 
0.81 / 0.81(\inc{0.09\%}) / 0.81(\inc{0.05\%}) & 
0.43 / 0.46(\inc{7.07\%}) / 0.46(\inc{7.92\%}) \\
Llama-3b & 
0.11 / 0.11(\inc{2.98\%}) / 0.11(\inc{2.68\%}) & 
0.81 / 0.81(\dec{0.09\%}) / 0.81(\inc{0.01\%}) & 
0.45 / 0.45(\dec{0.42\%}) / 0.45(\dec{1.27\%}) \\
Mistral-7b & 
0.11 / 0.12(\inc{7.77\%}) / 0.12(\inc{4.34\%}) & 
0.81 / 0.81(\inc{0.06\%}) / 0.81(\dec{0.01\%}) & 
0.45 / 0.48(\inc{9.22\%}) / 0.47(\inc{3.96\%}) \\
Qwen-7b & 
0.10 / 0.12(\inc{23.66\%}) / 0.15(\inc{50.51\%}) & 
0.81 / 0.81(\inc{0.21\%}) / 0.82(\inc{1.55\%}) & 
0.42 / 0.50(\inc{18.41\%}) / 0.54(\inc{27.77\%}) \\
Claude & 
0.09 / 0.17(\inc{102.23\%}) / 0.21(\inc{146.42\%}) & 
0.81 / 0.83(\inc{2.75\%}) / 0.84(\inc{3.29\%}) & 
0.37 / 0.54(\inc{45.86\%}) / 0.54(\inc{45.96\%}) \\
\hline
\multicolumn{4}{c}{\textbf{Interview Dataset}} \\
\hline
DeepSeek-8b & 
0.11 / 0.12(\inc{4.96\%}) / 0.12(\inc{5.33\%}) & 
0.81 / 0.82(\inc{0.43\%}) / 0.82(\inc{0.44\%}) & 
0.39 / 0.38(\dec{1.10\%}) / 0.38(\dec{3.42\%}) \\
Llama-3b & 
0.10 / 0.10(\dec{4.77\%}) / 0.10(\dec{2.97\%}) & 
0.81 / 0.81(\dec{0.19\%}) / 0.81(\dec{0.11\%}) & 
0.40 / 0.39(\dec{3.96\%}) / 0.39(\dec{3.36\%}) \\
Mistral-7b & 
0.11 / 0.11(\dec{7.96\%}) / 0.12(\inc{6.30\%}) & 
0.81 / 0.81(\dec{0.62\%}) / 0.82(\inc{0.41\%}) & 
0.41 / 0.40(\dec{1.86\%}) / 0.41(\dec{0.68\%}) \\
Qwen-7b & 
0.10 / 0.11(\inc{10.62\%}) / 0.11(\dec{2.42\%}) & 
0.81 / 0.81(\inc{0.22\%}) / 0.81(\inc{0.10\%}) & 
0.42 / 0.42(\inc{0.19\%}) / 0.43(\inc{1.44\%}) \\
Claude & 
0.12 / 0.15(\inc{32.90\%}) / 0.15(\inc{28.51\%}) & 
0.82 / 0.83(\inc{1.67\%}) / 0.83(\inc{1.54\%}) & 
0.43 / 0.49(\inc{13.36\%}) / 0.48(\inc{10.06\%}) \\
\bottomrule
\end{tabular}%
}
\vspace{-0.5em}
\caption{Overall result of the evaluation. Results are presented in \textit{Baseline / $DPRF_{Structured\_BAA}$ / $DPRF_{Free-Form\_BAA}$}. The percentage in parentheses is the relative improvement over baseline.}
\label{tab:performance_comparison}
\vspace{-1.5em}
\end{table*}
}

Our experiments demonstrate that \textbf{DPRF significantly improves the behavioral alignment of LLM RPAs with human ground truth} across a majority of tasks compared with a carefully curated baseline persona. 
More importantly, DPRF can consistently improve the persona over iterations in most scenarios.
We present the overall performance, analyze the effectiveness of the different cognitive analysis modules, and report the findings from our ablation study.

\subsection{Overall Performance}
As shown in Table \ref{tab:performance_comparison}, DPRF achieves consistent and substantial improvements over baseline across four of the five datasets: Debate, DepSeverity, IMDB, and CSSRS-Suicide. 
The refined personas lead to generated behaviors that are more aligned with the human ground truth.
Among the models, the most capable LLMs demonstrate the greatest improvements.
Specifically, Claude3.7-Sonnet exhibited the largest and most consistent performance gains, underscoring that a powerful LLM for the behavior analysis agent is key to maximizing the framework's effectiveness.
Qwen-2.5 (7B) and Mistral (7B) also realized steady improvements.
The performance of Llama-3.2 (8B) was more variable, occasionally showing decreased ROUGE-L despite gains in semantic similarity.

\subsection{High-Level Semantics vs. Lexical Fidelity}
A key insight from our results is that DPRF's impact is nuanced and task-dependent. We observe a distinction between its effect on \textbf{holistic semantic alignment} (measured by Sentence Embedding Similarity) and \textbf{fine-grained lexical fidelity} (measured by ROUGE-L and BERTScore).

In tasks like Opinionated Reviews and Mental Health Expression, which are often driven by abstract concepts (e.g., emotion or opinion), DPRF's primary contribution is enhancing the high-level core semantics. For instance, while smaller models showed limited gains in ROUGE-L, Claude 3.7-Sonnet achieved a remarkable 292.1\% increase in Sentence Embedding Similarity on the DepSeverity dataset. 
This result shows that DPRF helps grasp the correct meaning and intent, even if the precise wording doesn't match the ground truth.

On the other hand, in information-dense, reasoning-heavy tasks, like Formal Debates, DPRF significantly improves fine-grained lexical fidelity. For Qwen-2.5 (7B), ROUGE-L increased by 4.64\% while Sentence Embedding Similarity saw a more modest 1.70\% gain. 
The result suggests that for scenarios requiring the precise use of facts and arguments, DPRF helps the agent not only align its high-level stance but also select the correct keywords and phrasing.

\subsection{Structured vs. Free-Form Behavior Analysis}
A key finding is that the optimal behavioral analysis strategy is highly task-dependent and not having a "one-size-fits-all" solution, which hinges on the primary cognitive dimensions of the scenario.
Researchers deploying agent-based simulations should select their analysis method based on the cognitive demands of their target domain.

For \textbf{emotion-centric} tasks (i.e., DepSeverity, CSSRS-Suicide, IMDB), the simple free-form analysis agent consistently outperformed the theory-grounded structured agent. 
On the DepSeverity dataset, Claude 3.7-Sonnet with a free-form analysis yielded a 292.1\% improvement in Sentence Embedding Similarity, significantly higher than the 254.7\% gain from the structured ToM-driven analysis. 
The result implies that the structured ToM dimensions can sometimes over-constrain the analysis of nuanced, versatile emotional signals, whereas a free-form approach allows the agent to capture these characteristics more directly.

Conversely, for \textbf{cognitively complex} tasks requiring the integration of multiple dimensions (beliefs, goals, knowledge), the theory-grounded structured analysis was more effective. 
On the Debate task, Claude's ROUGE-L score improved by 27.7\% with the ToM-based analysis, compared to only 22.4\% with the free-form analysis. 
The ToM framework provides an essential scaffold for the agent to systematically dissect the different logical and intentional layers of the argument, leading to a more comprehensive and effective persona refinement.

\subsection{Ablation: Persona Profile in Behavior Analysis Agent}

To assess whether persona profile is necessary in the behavior analysis, we conduct an ablation study on Claude3.7-Sonnet: (i) a no-persona version that receives only the generated response, ground truth, and background content, and (ii) a persona version that additionally receives the current persona description. 
For each dataset, we randomly sample 100 examples and run both variants under identical settings.
The results, detailed in Appendix \ref{fig:claude_structured_comparison}, show that the \textbf{"with-persona" variant consistently outperforms the "no-persona" variant} across nearly all metrics and datasets.

The improvement is particularly pronounced on the DepSeverity and CSSRS-Suicide datasets.
This finding confirms a core hypothesis of our work: the persona serves as a critical anchor for the analysis. 
Without it, the agent can only perform a context-free comparison of two text-based behavior descriptions; however, with the input of the persona, the agent can assess RPA's behavior with respect to the intended identity and achieve a more targeted and effective analysis of cognitive divergences.

\subsection{Boundary Condition: The PublicInterview Challenge}

The DPRF framework's performance was often limited on our new PublicInterview dataset. 
We attribute this not to a failure of the framework itself, but to the inherent substantial complexity of the task. 
Interview responses are governed by a rich set of situated factors beyond a static persona, such as the environmental context, interview's topic, interviewer's style, and the public figure's immediate strategic goals.
This result is particularly valuable because it \textbf{identifies a potential boundary condition for current persona-based RPAs}. 
Capturing such highly dynamic and context-dependent behaviors may require future agent architectures that can integrate persona profiles with real-time environmental and social cues. 
Thus, the PublicInterview dataset serves as an important and challenging benchmark for the next generation of LLM RPAs.

\section{Conclusion and Future Work}

This work introduced the \textbf{Dynamic Persona Refinement Framework (DPRF)}, a novel methodology to improve the behavioral fidelity of LLM agents by moving beyond static, manually created persona profiles to a data-driven optimization problem.
DPRF allows iterative analysis of cognitive divergences against human ground truth and refining the persona to address the divergences.
Our evaluation across four diverse scenarios with five different LLMs confirms that DPRF consistently enhances behavioral alignment, and the effectiveness is generalizable across models and scenarios.

Our research yields several key insights. First, we demonstrated that the nature of improvement is task-dependent: DPRF enhances high-level semantic meaning in emotionally-driven tasks and fine-grained lexical fidelity in information-dense, logical tasks. 
Second, the optimal behavioral analysis strategy depends on the task's cognitive complexity, with free-form analysis excelling in emotion-centric domains and a theory-grounded (ToM) structured analysis proving superior for multi-dimensional reasoning tasks.

The implications of this work are significant. 
By providing a systematic process for persona refinement and validation, DPRF addresses a critical gap in the development of reliable agent-based simulations for social science research, user experience testing, and multi-perspective evaluation systems. 
It lays the groundwork for creating truly personalized AI assistants that can dynamically and continuously adapt to their users' unique behaviors, preferences, and cognitive styles, moving beyond one-shot personalization.

\section{Limitations}

While our findings establish DPRF as a robust method for persona refinement, it's important to situate this work within its specific methodological context and acknowledge its boundaries. We outline three key areas for consideration and future exploration.
First, our evaluation was intentionally designed around four distinct scenarios, each chosen to probe a different core cognitive activity: logical reasoning (Debates), emotional expression (Mental Health), opinion formation (Reviews), and goal-oriented conversation (Interviews). 
While this provides a strong foundational proof-of-concept, these text-based tasks represent only a subset of complex, real-world human behavior. 
The performance of DPRF in more dynamic, interactive, or multi-modal environments remains an open question. 
Future work should therefore focus on testing the framework's generalizability in richer contexts, such as multi-turn conversational agents, collaborative task simulations, or even scenarios where the ground truth includes non-textual cues.

Second, establishing a fair performance baseline is a known challenge in persona-based generation, particularly for datasets that lack standardized persona profiles. 
For consistency, we constructed our own baseline personas where necessary.
Nevertheless, the central contribution of DPRF is a consistent process of iterative improvement.
DPRF's value lies in its ability to take a generic persona and systematically specialize it against behavioral evidence with high interpretability.

Third, our experiments position DPRF as a gradient-free, inference-time optimization framework. 
We did not compare it directly with training-based approaches like fine-tuning on target data.
In particular, we see DPRF not as a replacement for fine-tuning, but as a complementary methodology because of its distinct advantages of data and computational efficiency, as well as interpretability.
A systematic comparison investigating the trade-offs between diverse approaches remains a promising direction for future research.

\bibliography{anthology,custom}

\appendix
\clearpage
\section{Ablation result}
\label{app:ablation-result}
\begin{figure*}[!b]
    \centering

    \begin{subfigure}[b]{0.31\linewidth}
        \centering
        \includegraphics[width=\linewidth,trim={0 .4cm 0 0},clip]{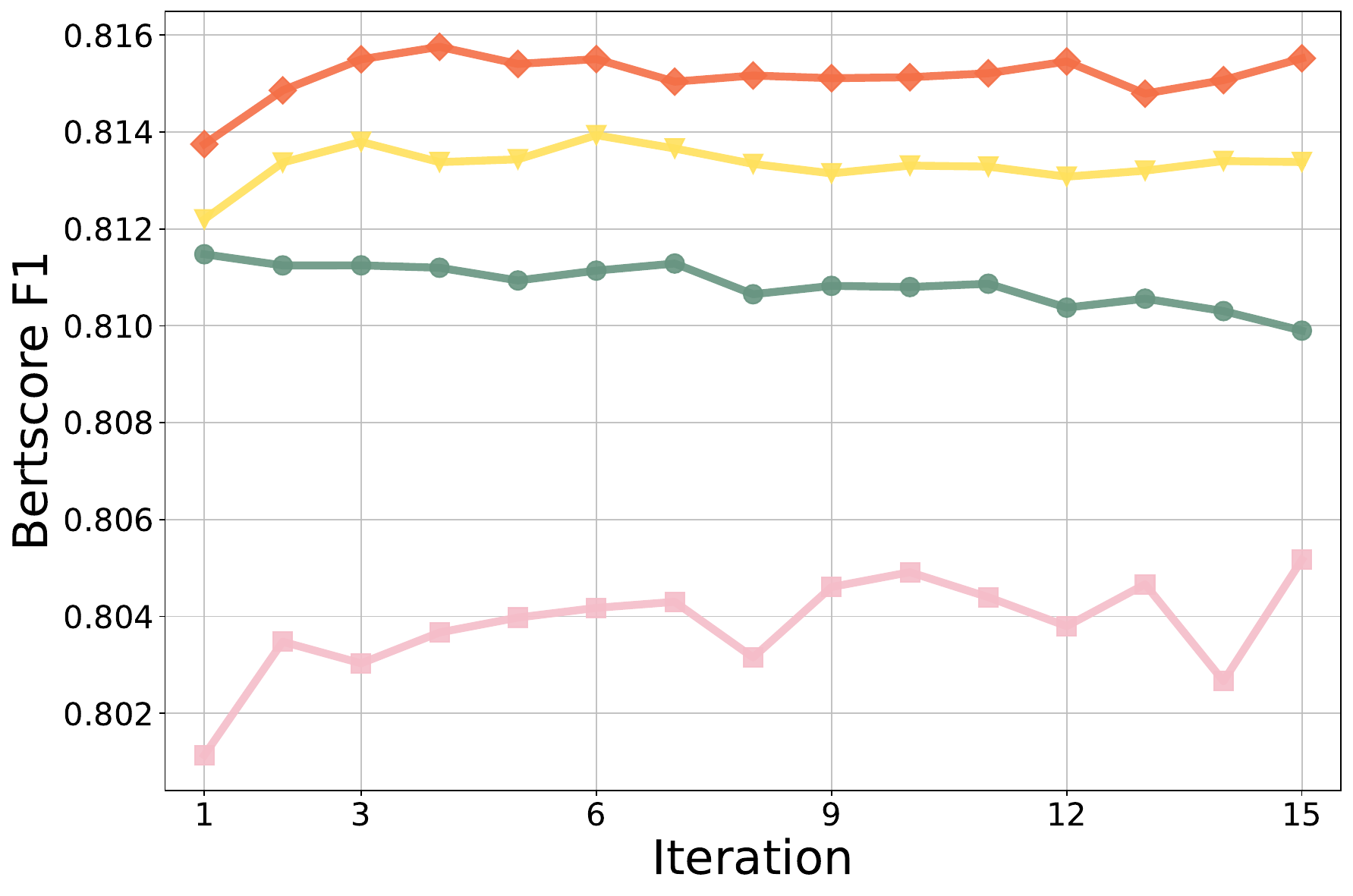}
        \caption{Debate}
        \label{fig:bert-debate}
    \end{subfigure}\hfill
    \begin{subfigure}[b]{0.31\linewidth}
        \centering
        \includegraphics[width=\linewidth,trim={0 .4cm 0 0},clip]{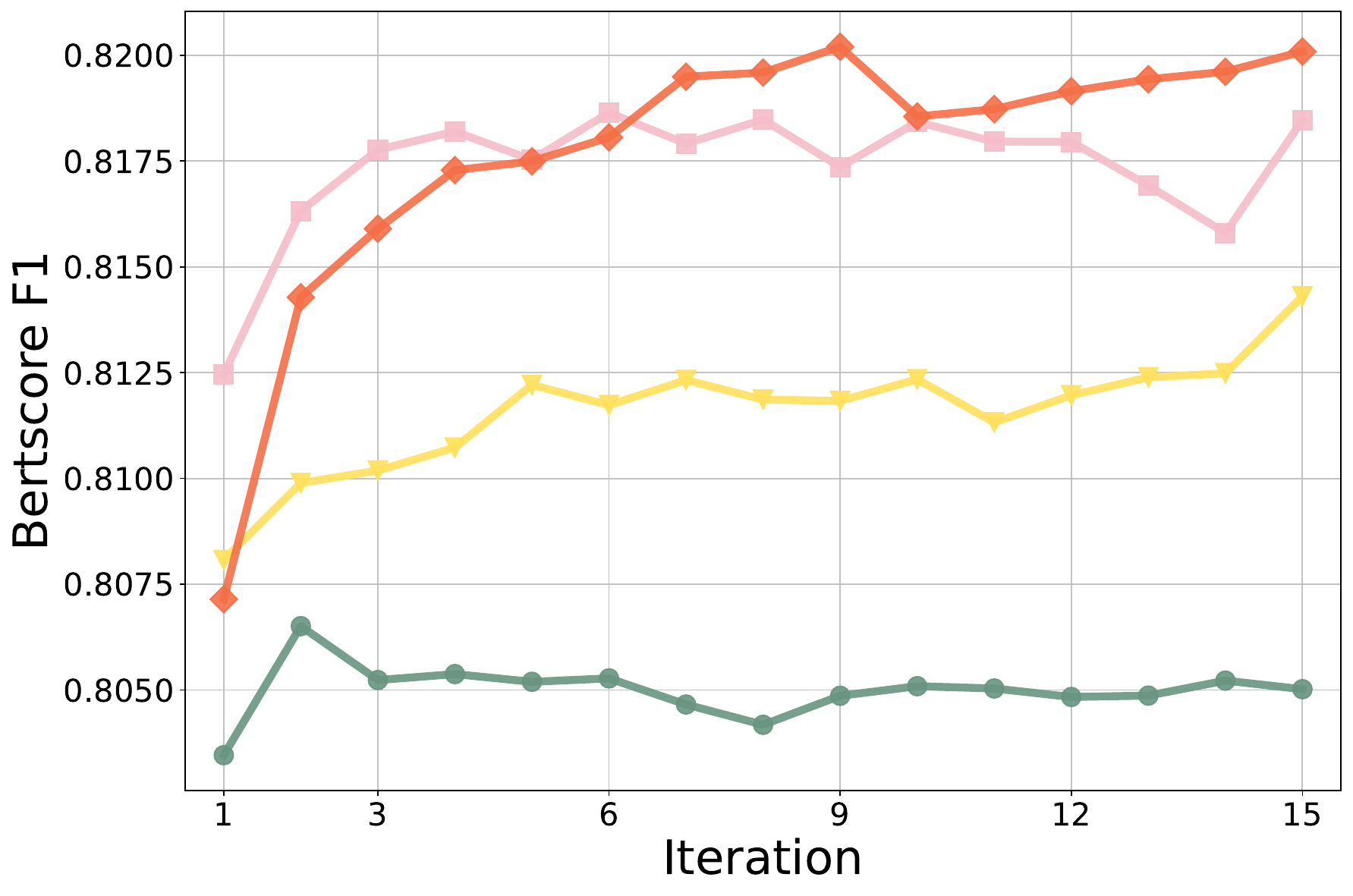}
        \caption{Depression}
        \label{fig:bert-depression}
    \end{subfigure}\hfill
    \begin{subfigure}[b]{0.31\linewidth}
        \centering
        \includegraphics[width=\linewidth,trim={0 .4cm 0 0},clip]{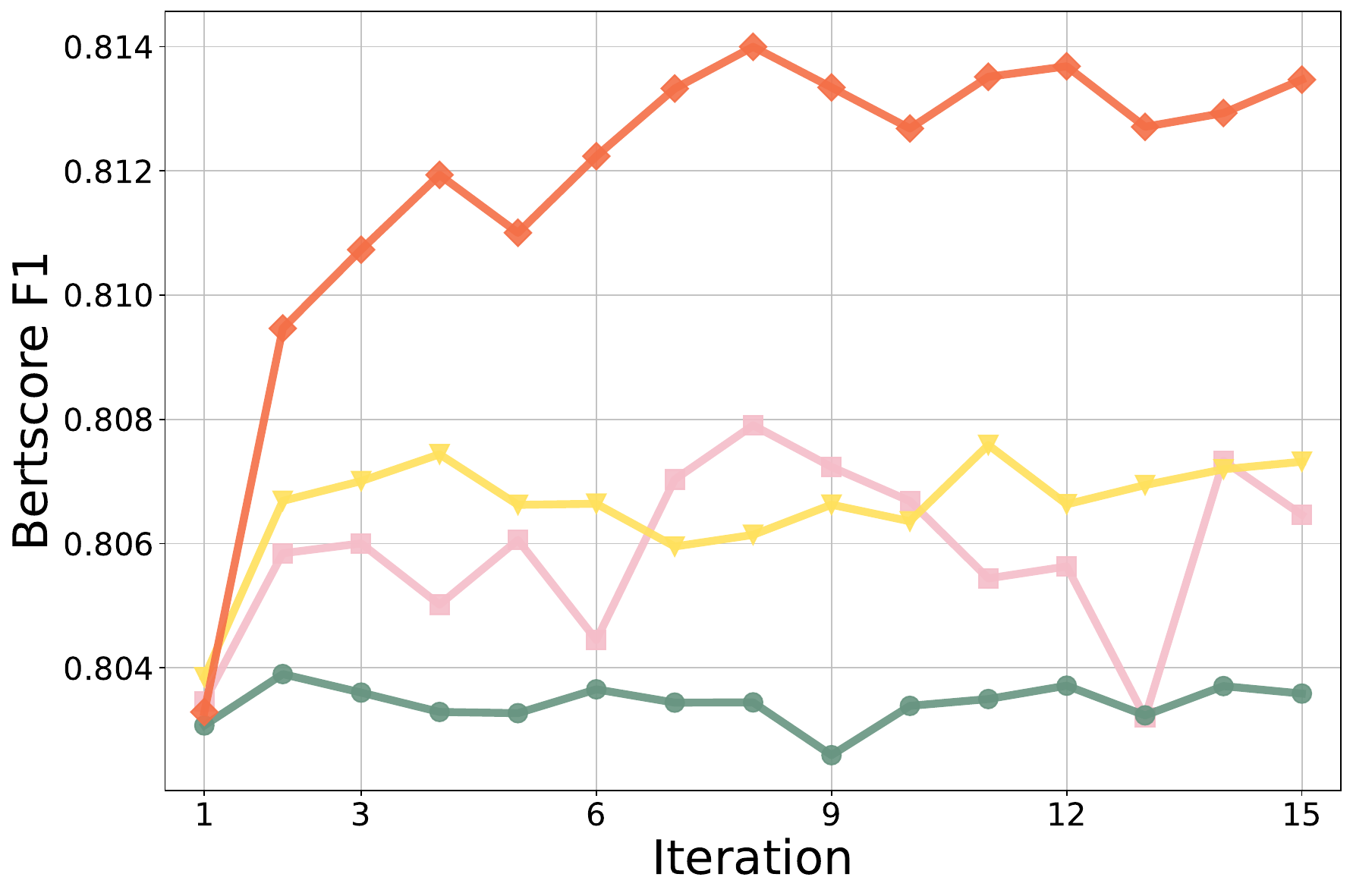}
        \caption{IMDB}
        \label{fig:bert-imdb}
    \end{subfigure}

    \vspace{6pt} %
    \begin{subfigure}[b]{0.31\linewidth}
        \centering
        \includegraphics[width=\linewidth,trim={0 .4cm 0 0},clip]{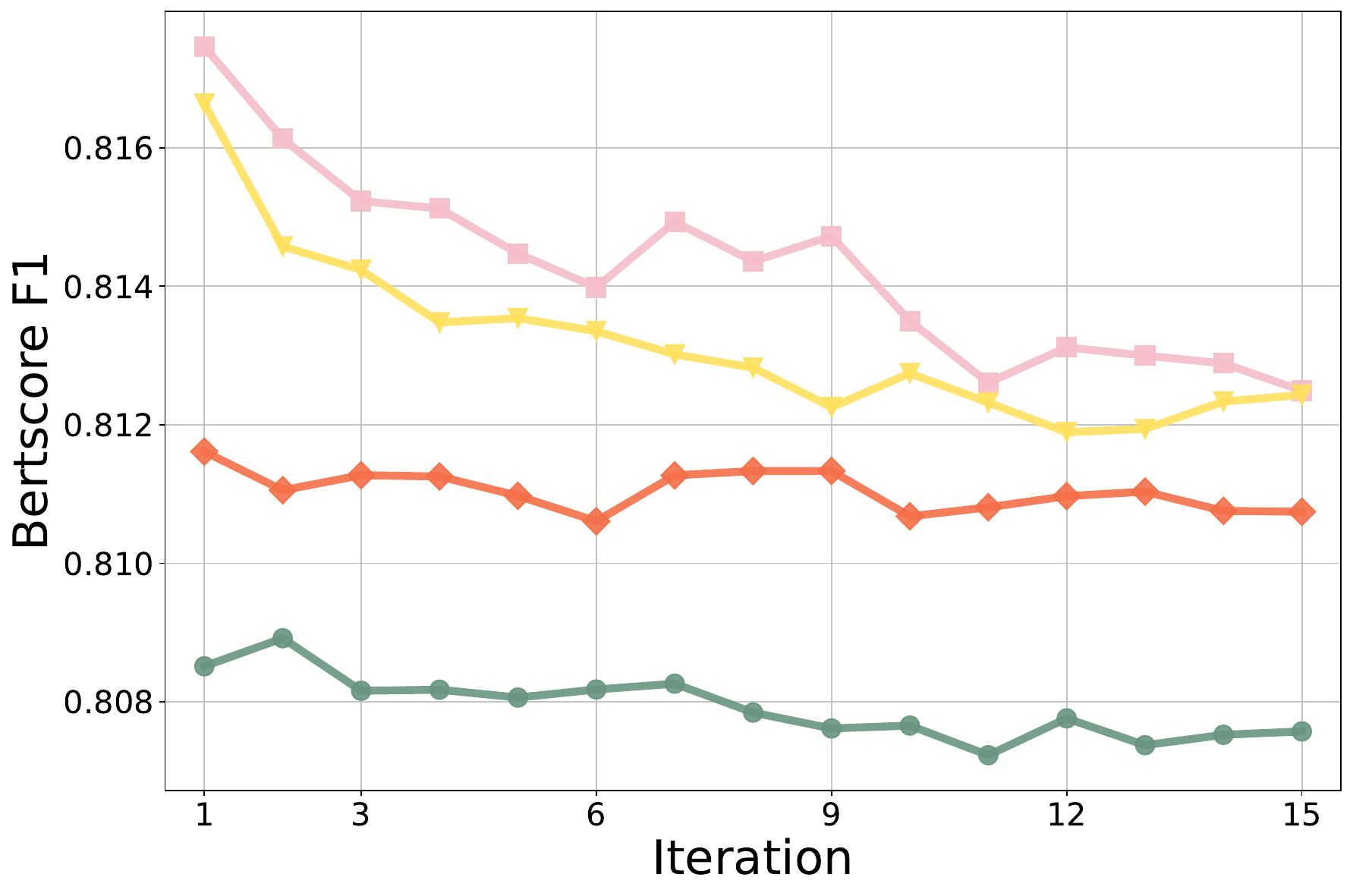}
        \caption{Interview}
        \label{fig:bert-interview}
    \end{subfigure}\hfill
    \begin{subfigure}[b]{0.31\linewidth}
        \centering
        \includegraphics[width=\linewidth,trim={0 .4cm 0 0},clip]{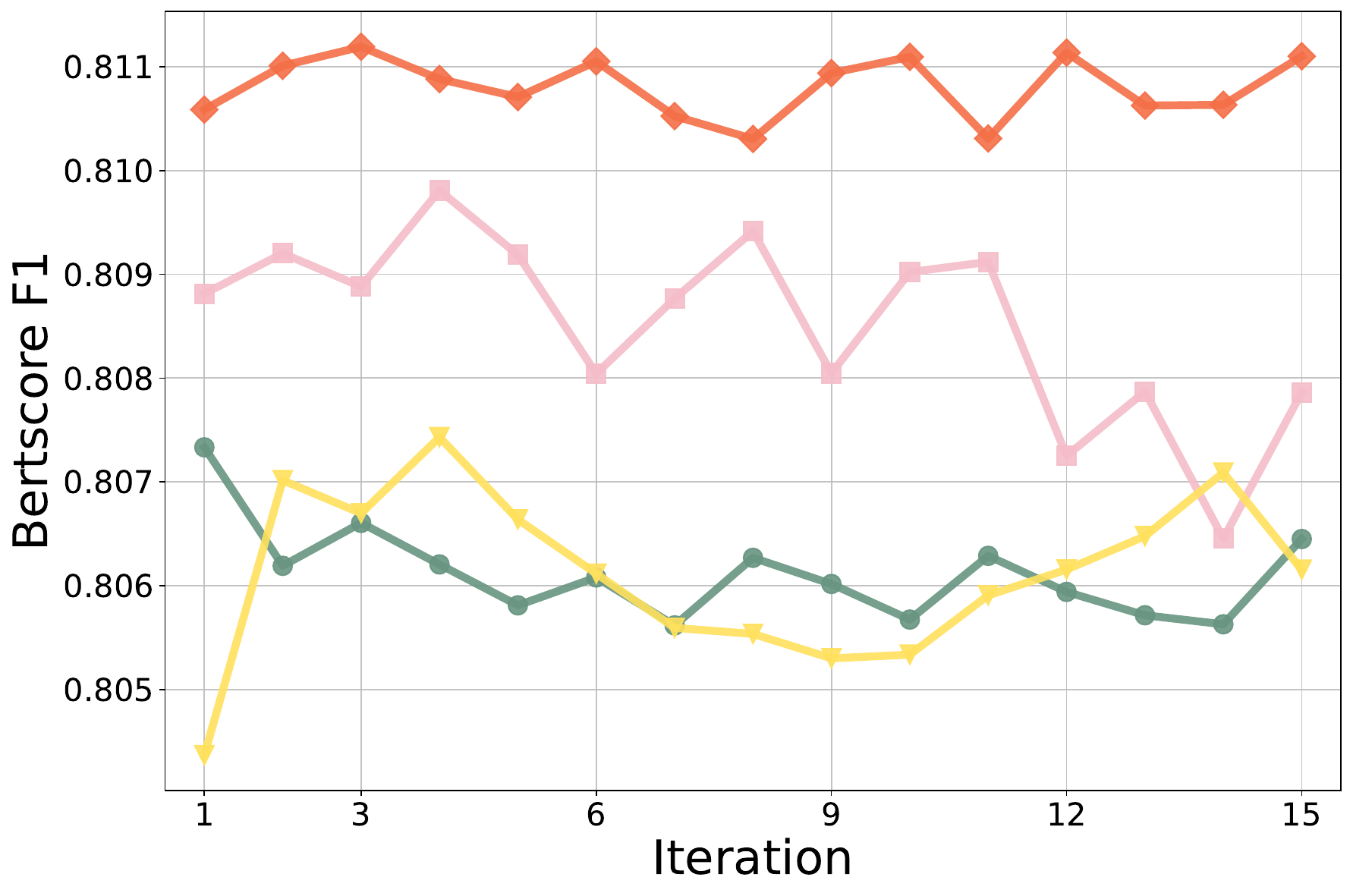}
        \caption{Suicide}
        \label{fig:bert-suicide}
    \end{subfigure}\hfill
    \begin{subfigure}[b]{0.31\linewidth}
        \centering
        \raisebox{0.6cm}{%
            \includegraphics[width=0.70\linewidth,trim={0 .4cm 0 0},clip]{resources/picture/models_legend.pdf}%
        }
    \end{subfigure}
    \caption{Bertscore-f1 on small models across different datasets}
    \label{fig:bertscore-comparison}
\end{figure*}

\begin{figure*}[!b]
    \centering

    \begin{subfigure}[b]{0.31\linewidth}
        \centering
        \includegraphics[width=\linewidth,trim={0 .4cm 0 0},clip]{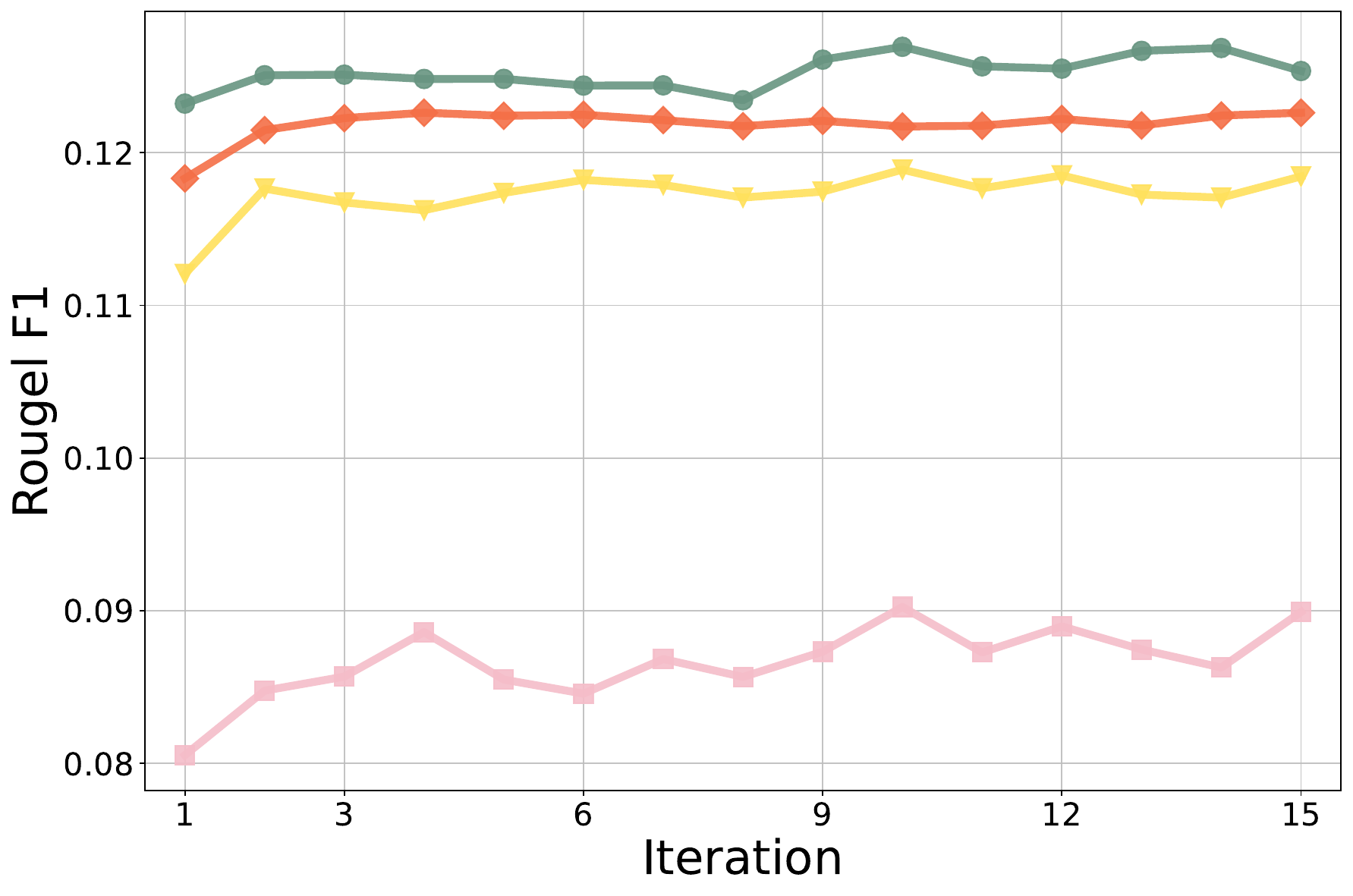}
        \caption{Debate}
        \label{fig:rougel-debate}
    \end{subfigure}\hfill
    \begin{subfigure}[b]{0.31\linewidth}
        \centering
        \includegraphics[width=\linewidth,trim={0 .4cm 0 0},clip]{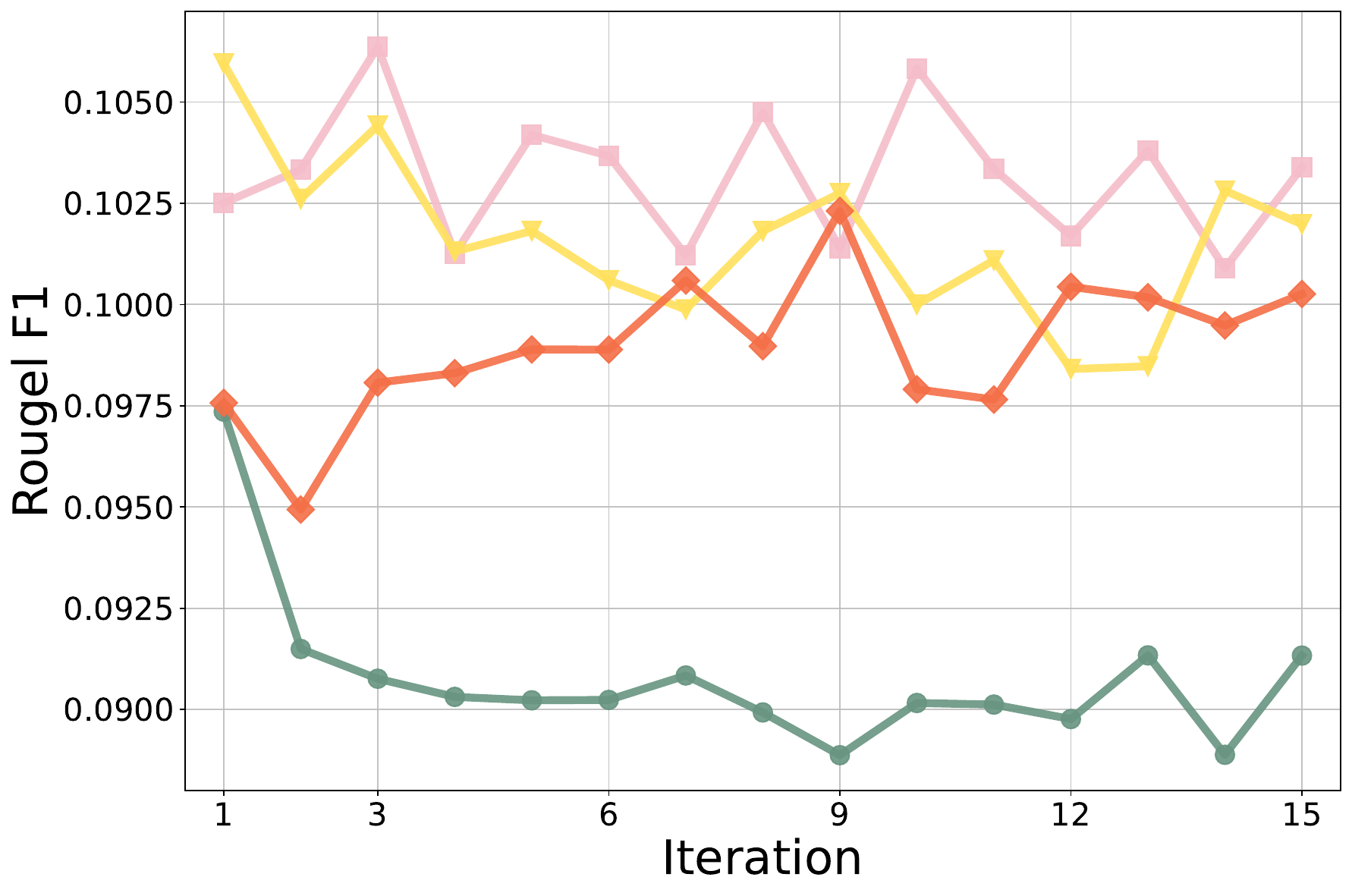}
        \caption{Depression}
        \label{fig:rougel-depression}
    \end{subfigure}\hfill
    \begin{subfigure}[b]{0.31\linewidth}
        \centering
        \includegraphics[width=\linewidth,trim={0 .4cm 0 0},clip]{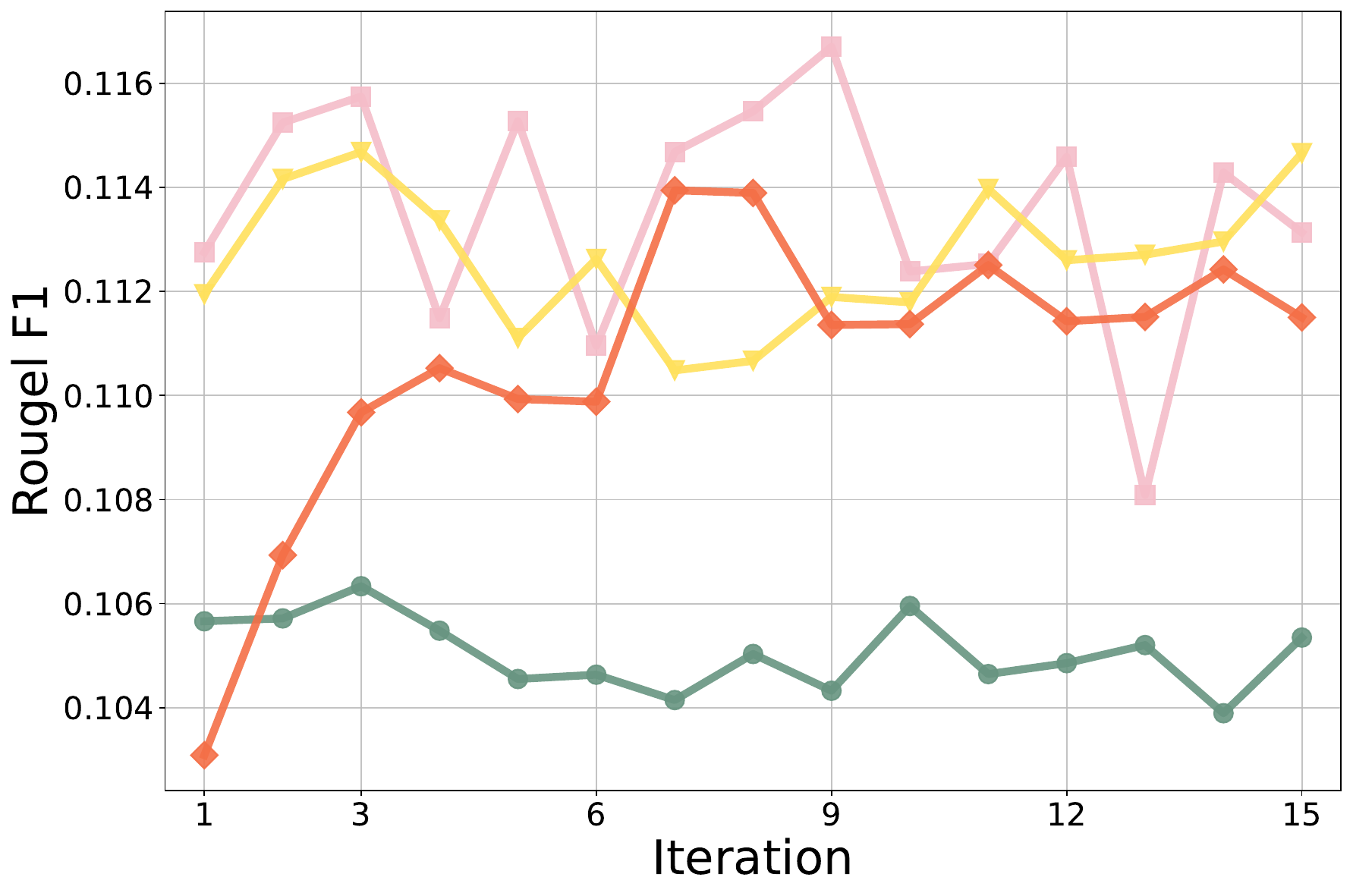}
        \caption{IMDB}
        \label{fig:rougel-imdb}
    \end{subfigure}

    \vspace{6pt} %
    \begin{subfigure}[b]{0.31\linewidth}
        \centering
        \includegraphics[width=\linewidth,trim={0 .4cm 0 0},clip]{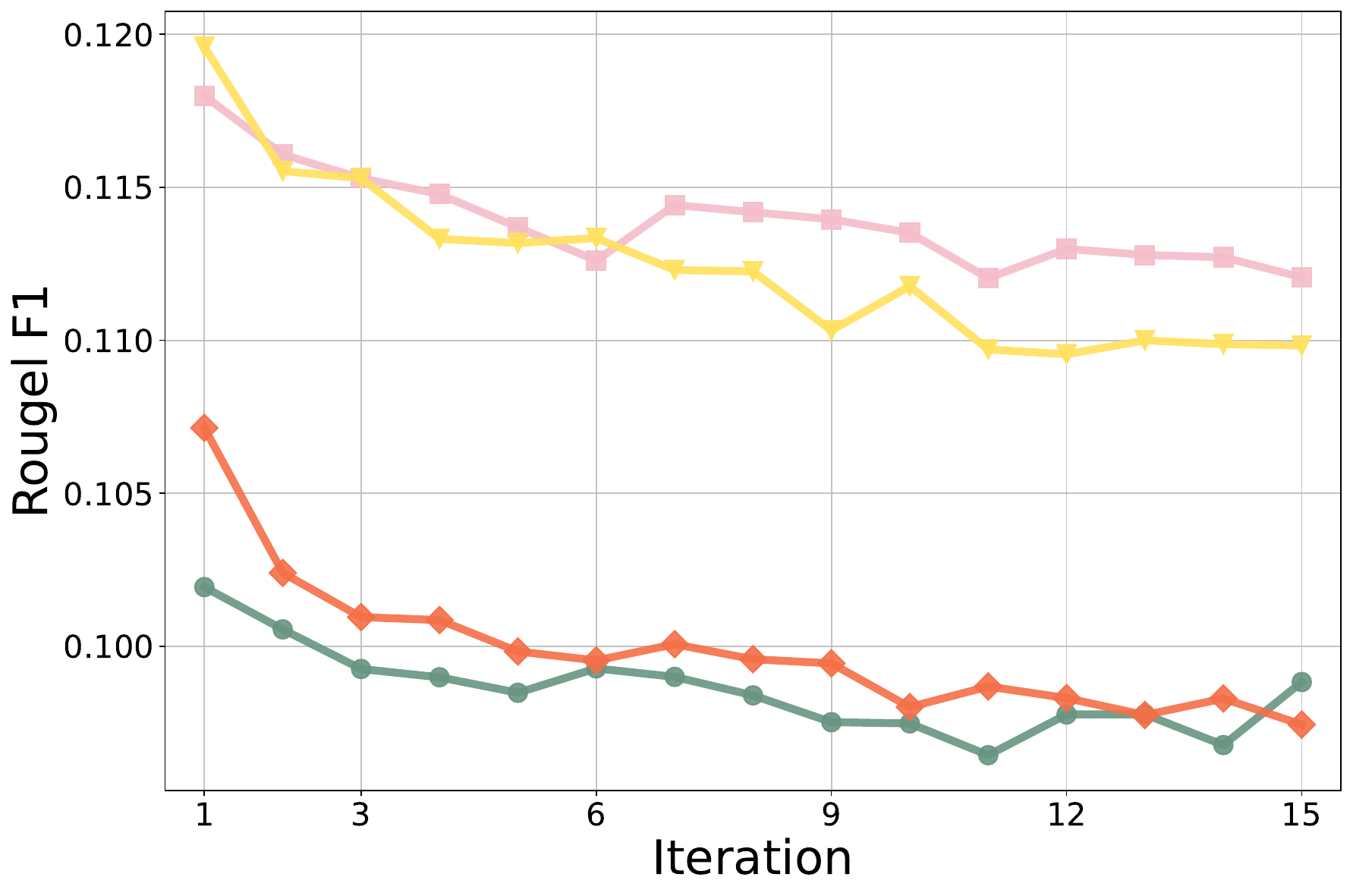}
        \caption{Interview}
        \label{fig:rougel-interview}
    \end{subfigure}\hfill
    \begin{subfigure}[b]{0.31\linewidth}
        \centering
        \includegraphics[width=\linewidth,trim={0 .4cm 0 0},clip]{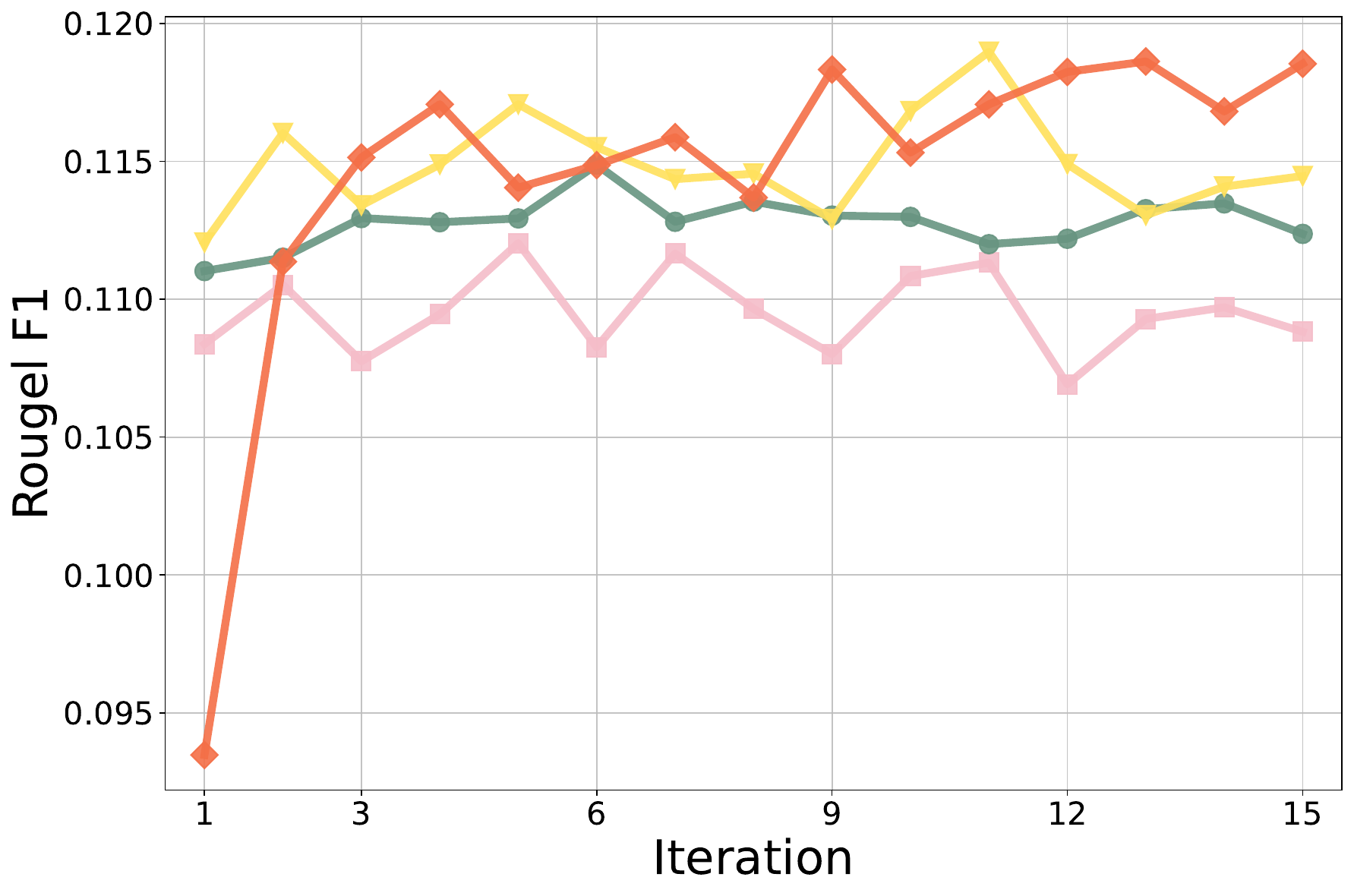}
        \caption{Suicide}
        \label{fig:rougel-suicide}
    \end{subfigure}\hfill
    \begin{subfigure}[b]{0.31\linewidth}
        \centering
        \raisebox{0.6cm}{%
            \includegraphics[width=0.70\linewidth,trim={0 .4cm 0 0},clip]{resources/picture/models_legend.pdf}%
        }
    \end{subfigure}
    \caption{RougeL-f1 on small models across different datasets}
    \label{fig:rougel_comparison}
\end{figure*}

\begin{figure*}[htbp]
    \centering

    \begin{subfigure}[b]{0.31\linewidth}
        \centering
        \includegraphics[width=\linewidth,trim={0 .4cm 0 0},clip]{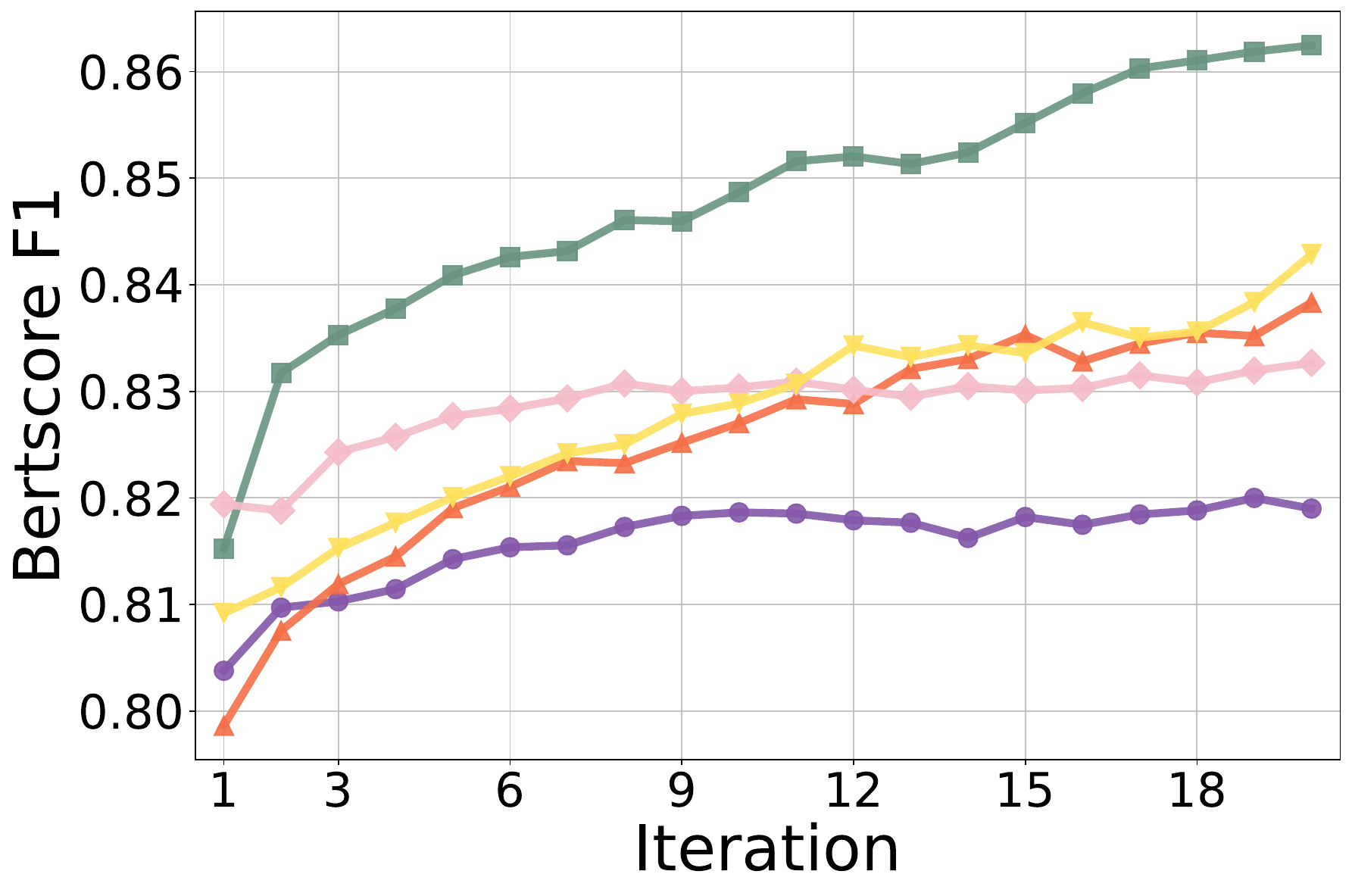}
        \label{fig:claude-bert_structured}
    \end{subfigure}\hfill
    \begin{subfigure}[b]{0.31\linewidth}
        \centering
        \includegraphics[width=\linewidth,trim={0 .4cm 0 0},clip]{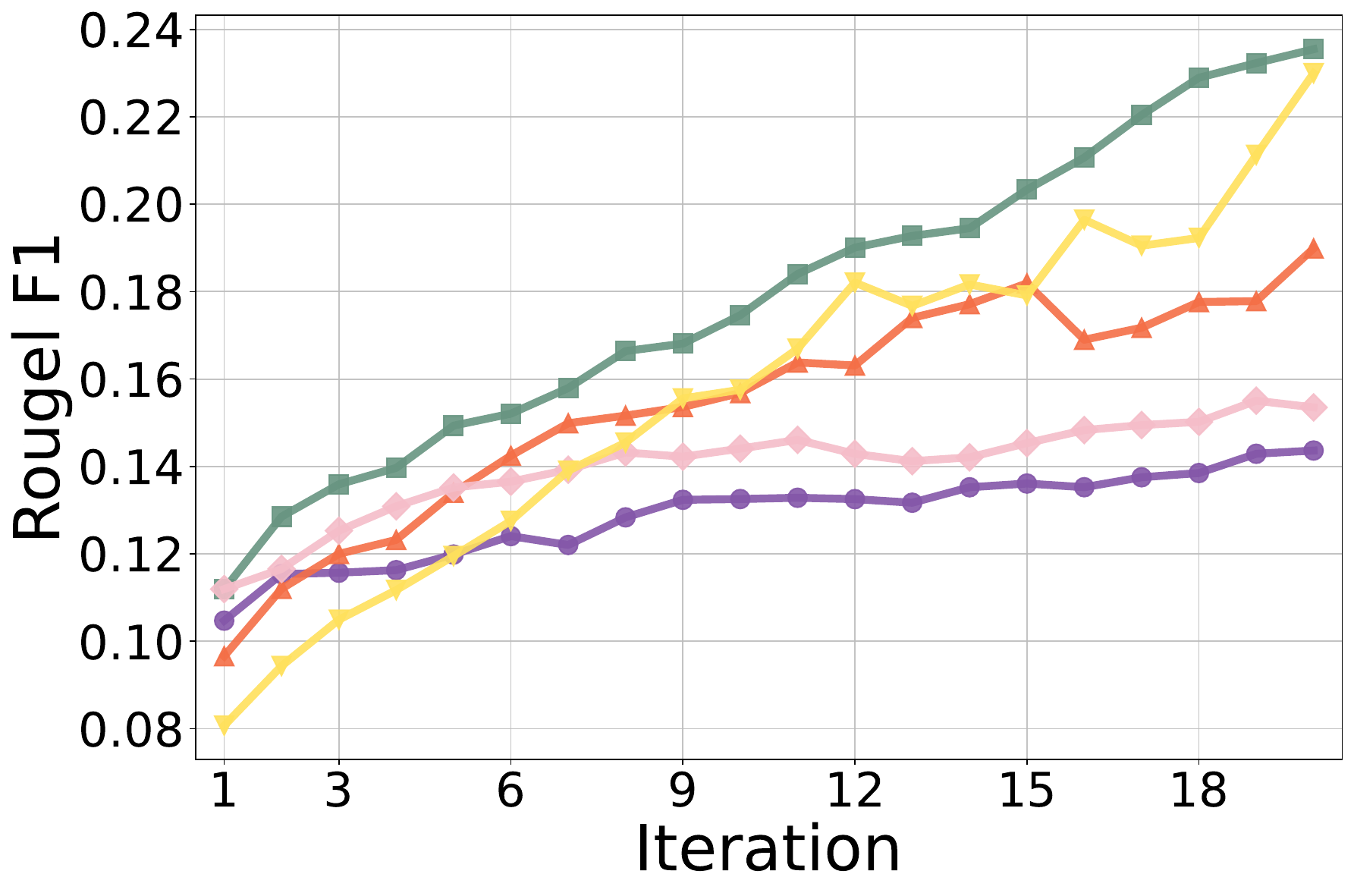}
        \label{fig:claude_rougel_structured}
    \end{subfigure}\hfill
    \begin{subfigure}[b]{0.31\linewidth}
        \centering
        \includegraphics[width=\linewidth,trim={0 .4cm 0 0},clip]{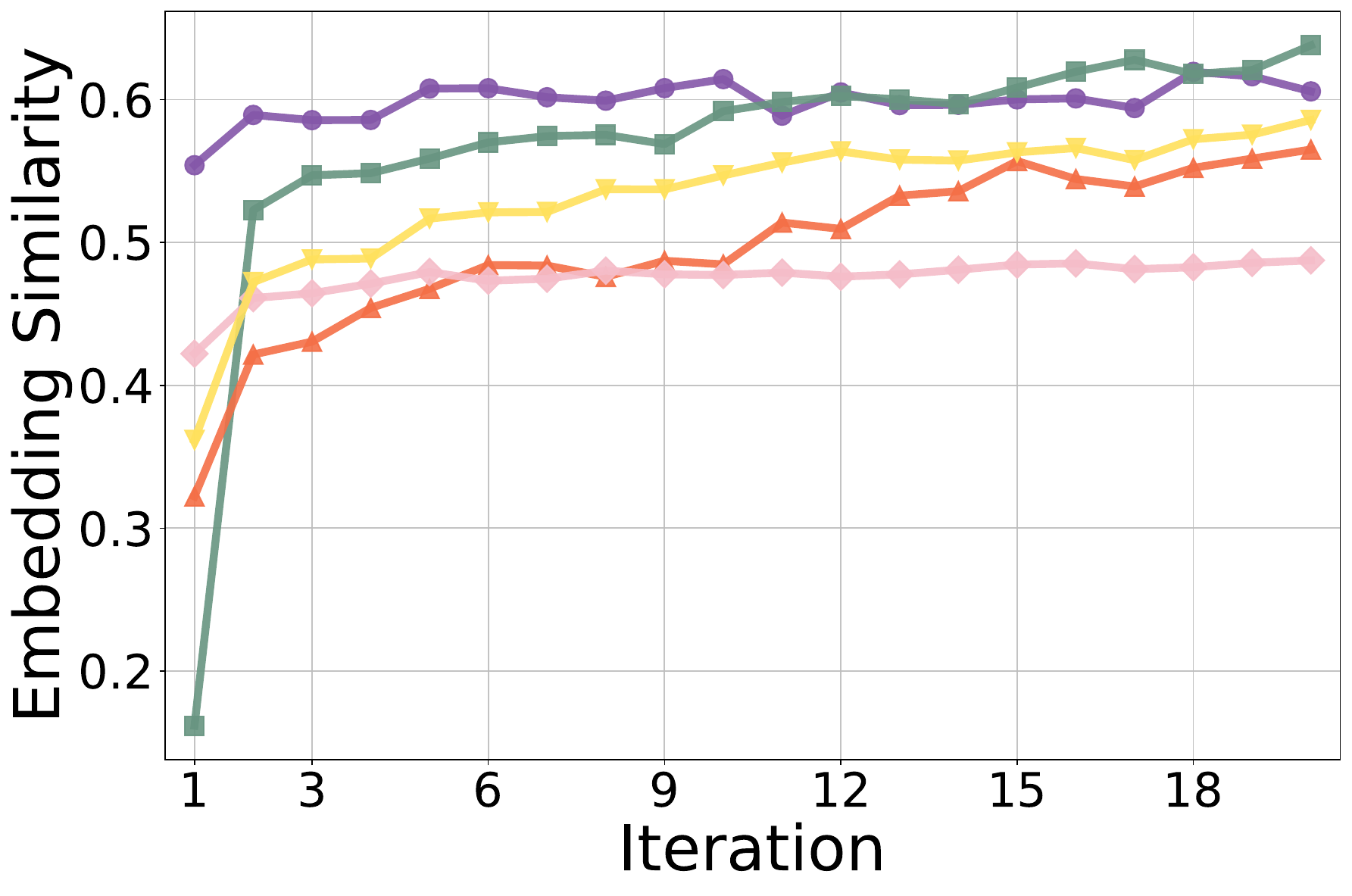}
        \label{fig:claude_embedding_structured}
    
    \end{subfigure}
        
    \vspace{6pt} %
    \begin{subfigure}[b]{0.31\linewidth}
        \centering
        \raisebox{0.6cm}{%
            \includegraphics[width=0.70\linewidth,trim={0 .4cm 0 0},clip]{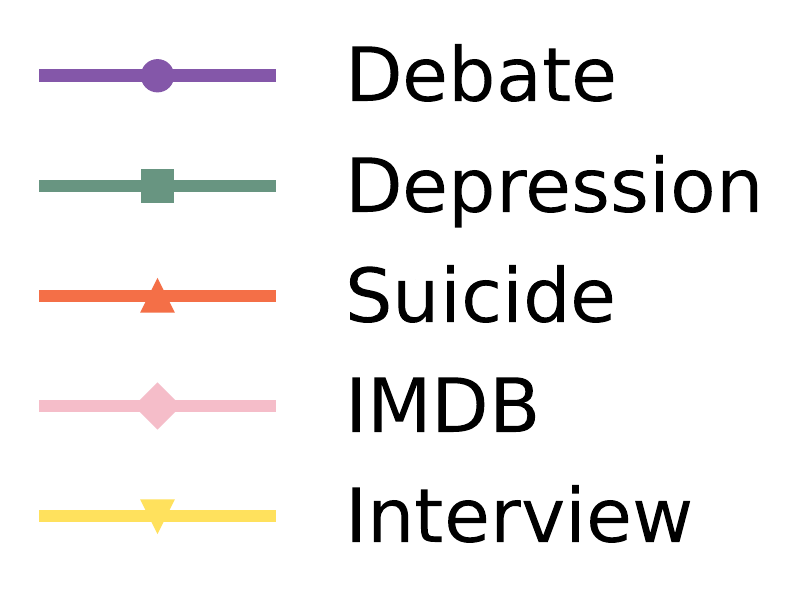}%
        }
    \end{subfigure}        
    \caption{Claude ablation result with structured analysis prompt}
    \label{fig:claude_structured_comparison}
    
\end{figure*}

\begin{figure*}[htbp]
    \centering

    \begin{subfigure}[b]{0.31\linewidth}
        \centering
        \includegraphics[width=\linewidth,trim={0 .4cm 0 0},clip]{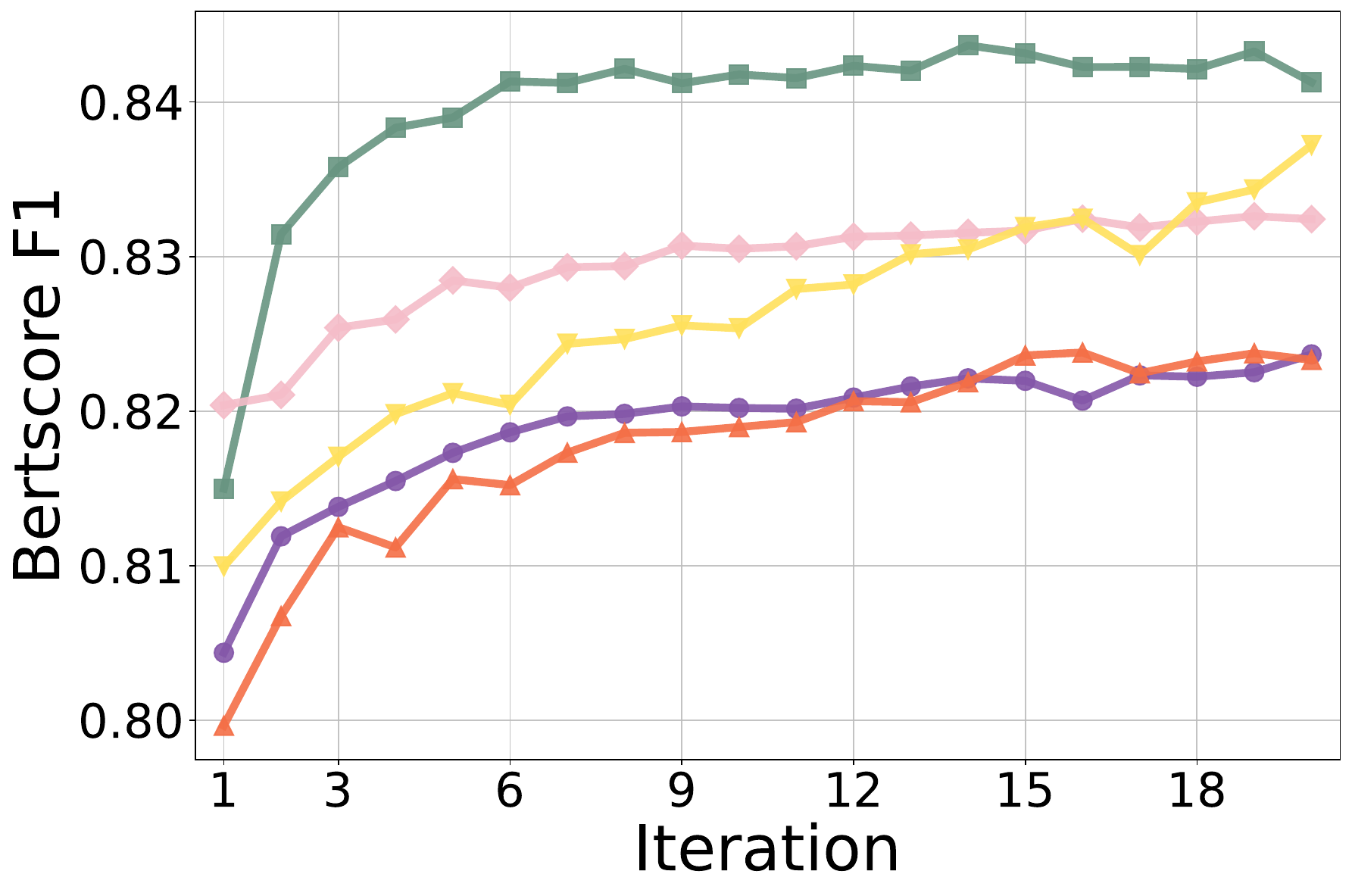}
        \label{fig:claude_bert_nopersona}
    \end{subfigure}\hfill
    \begin{subfigure}[b]{0.31\linewidth}
        \centering
        \includegraphics[width=\linewidth,trim={0 .4cm 0 0},clip]{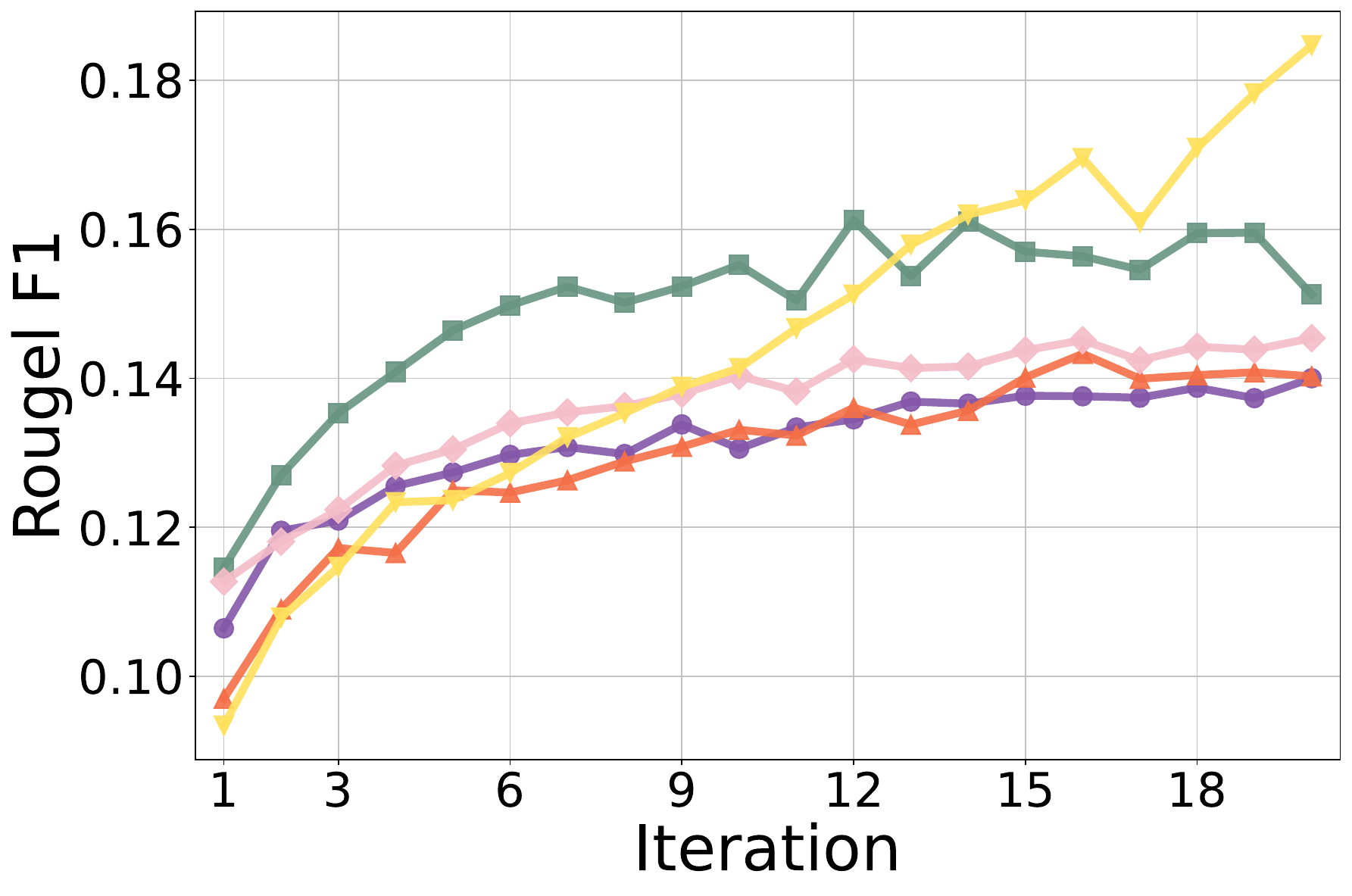}
        \label{fig:claude_rougel_nopersona}
    \end{subfigure}\hfill
    \begin{subfigure}[b]{0.31\linewidth}
        \centering
        \includegraphics[width=\linewidth,trim={0 .4cm 0 0},clip]{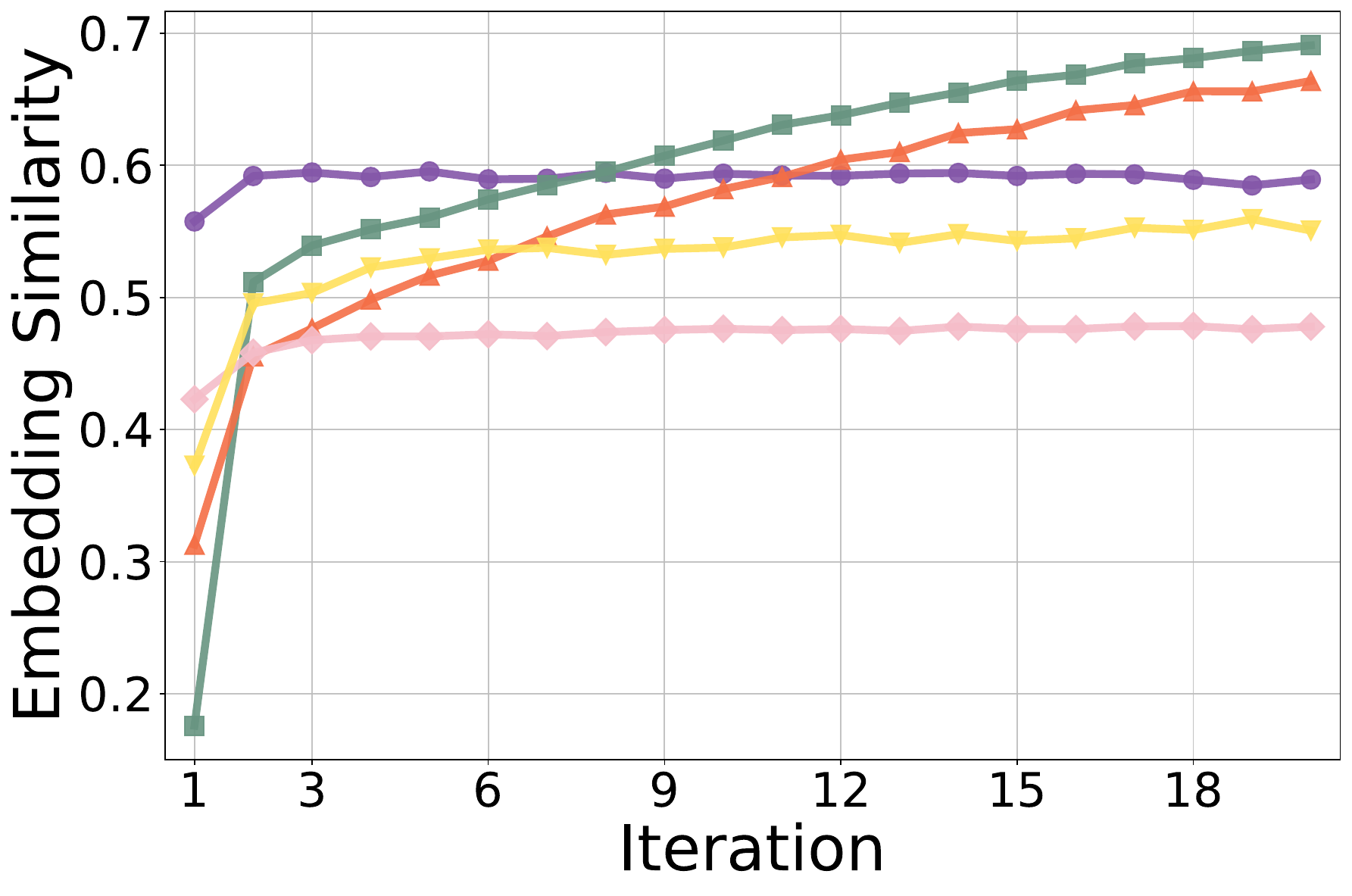}
        \label{fig:claude_embedding_nopersona}
    \end{subfigure}
    
    \vspace{6pt} %
    \begin{subfigure}[b]{0.31\linewidth}
        \centering
        \raisebox{0.6cm}{%
            \includegraphics[width=0.70\linewidth,trim={0 .4cm 0 0},clip]{resources/picture/claude_datasets_legend.pdf}%
        }
    \end{subfigure}
    \caption{Claude ablation result with no-persona analysis prompt}
    \label{fig:claude_nopersona_comparison}
\end{figure*}

\nopagebreak[4]
\FloatBarrier
In this section, we present the remaining ablation results: BERTScore‑F1 and ROUGE‑L F1 for the small models, and BERTScore‑F1, ROUGE‑L F1, and Embedding Similarity for the Claude model. Figure \ref{fig:bertscore-comparison} and \ref{fig:rougel_comparison} compares the performance of different small model on one metric in a plot. While in figure \ref{fig:claude_structured_comparison} and \ref{fig:claude_nopersona_comparison} shows result of Claude model, and we compare the performance of different dataset on one metric in a plot.

\section{Experimental Hyperparameters and Configuration Details}
\label{app:exp-config}

This section provides comprehensive details of the hyperparameters and experimental configurations used across all experiments in this work. 

\subsection{Claude Experimental Configuration}\label{app:base-config}
We use Amazon Bedrock to generate Claude responses. To improve robustness, we set the maximum retry count to 100 in case of connection failures. All experiments follow a consistent base configuration with the following shared hyperparameters:\\

\begin{tabular}{@{} p{.30\linewidth} p{.59\linewidth} @{}}
\textbf{Model} & \seqsplit{us.anthropic.claude-3-7-sonnet-20250219-v1:0} \\
\textbf{Refinement iterations} & 20 \\
\textbf{Temperature} & 0.6 \\
\textbf{Max tokens} & 2000 \\
\textbf{Top-p} & 0.95 \\
\textbf{Max attempts} & 100 \\
\textbf{AWS Region} & us-east-1 \\
\end{tabular}

\subsection{Open-Source Model Experiments}\label{app:opensource}
We evaluate four open‑source models using the sglang framework, and we list their corresponding Hugging Face model names as follows.
\par\medskip
\begin{tabular}{@{} p{.30\linewidth} p{.59\linewidth} @{}}
\textbf{DeepSeek-R1} & \seqsplit{deepseek-ai/DeepSeek-R1-Distill-Llama-8B} \\
\textbf{Llama-3.2}   & \seqsplit{meta-llama/Llama-3.2-3B-Instruct} \\
\textbf{Mistral}     & \seqsplit{mistralai/Mistral-7B-Instruct-v0.3} \\
\textbf{Qwen}        & \seqsplit{Qwen/Qwen2.5-7B-Instruct} \\
\end{tabular}

\section{Data Collection Details of PublicInterview Dataset}
\label{interview_process}
Our data collection process for the PublicInterview dataset began with 3,361 entries from the Government and Business sections of the Personality Database (PDB). 
We first filtered this pool by removing individuals who died before 1980, due to the limited availability of interview content on YouTube, and those with ambiguous MBTI labels (marked as 'XXXX' on the website). 
This step reduced the number of potential candidates to 2,110. 
To ensure the reliability of the personality labels, we further refined the list by retaining only profiles with more than three MBTI classification votes, which resulted in a final set of 1,014 candidate profiles for data collection. 

For each candidate, we conducted a targeted YouTube search using the query ``\texttt{"\{candidate's name\} interview"}'' and collected metadata from the top 10 results, including video titles, descriptions, and subtitles.
We then employed a Claude-based validation process to confirm that each video was indeed about the target individual. 
Subsequently, we applied language filtering to exclude non-English content, thereby preventing potential semantic distortions from translation. 
This two-stage validation process yielded interviews for 772 individuals.

For audio processing, we employed pyannote-audio for precise speaker diarization to obtain timestamps and identify distinct speakers, followed by high-quality transcription of the audio segments using Whisper. 
We then utilized Claude 3.5 to identify which speaker in the transcript corresponded to the target candidate. 
We excluded interviews where the model could not confidently identify the candidate or where the total number of conversational turns was less than eight, resulting in a set of interviews from 605 public figures. 
From the transcripts of these interviews, we extracted analysis segments. 
For a segment to be included, we required that in the preceding two conversational turns, both the interviewee and the interviewer must have spoken at least 600 characters, and the celebrity's own response must contain at least 50 characters. 
This final selection process yielded 2,820 segments covering 564 distinct public figures.

\section{Analysis Prompts}
Table \ref{tab:prompt_brief}, \ref{tab:prompt_structured} and \ref{tab:prompt_nopersona} present three prompt variants used in the Behavior Analysis Agent. The free-form prompt serves as a baseline that satisfies the task requirements. The structured prompt integrates dimensions from Theory of Mind. The no‑persona prompt is derived from the structured version by removing the persona component.

\subsection{Free-Form Analysis}

Table \ref{tab:prompt_brief} shows the free-form prompt we use in Behavior Analysis Agent.

\begin{table*}[!ht]
\centering
\begin{booktabs}{
    width=\linewidth,
    colspec={Q[l,co=1,m]},
    cell{1}{1}={c},
}
\toprule
{\textbf{Prompt for Analysis 1 : Free-Form Analysis}}\\
\midrule
\midrule
You are an expert in cognitive science. Your task is to analyze the cognitive differences between the current response based on the persona description and the expected response of a target human.\\
\\
PERSONA:\\
\{persona\}\\
\\
BACKGROUND INFORMATION:\\
\{content\}\\
\\
CURRENT RESPONSE:\\
\{generated\_response\}\\
\\
EXPECTED RESPONSE:\\
\{ground\_truth\}\\
\\
Conduct a comprehensive, structured analysis comparing the current behavioral/non-behavioral responses to the ideal one, and explicitly state what should be incorporated into the persona description to produce individualized behavioral/non-behavioral responses that more closely match the ideal one.\\
\\
Your response should provide analysis conclusions with reasons and specific examples (if applicable), formatted as numbered points: 1. 2. 3. etc.\\
\bottomrule
\end{booktabs}
\caption{Baseline setting for analyzing differences between generated responses and ground truth.}
\label{tab:prompt_brief}
\end{table*}

\subsection{Theory-Grounded Structured Analysis}
Table \ref{tab:prompt_structured} shows theory-grounded structured prompt we use in Behavior Analysis Agent.
\begin{table*}[!ht]
\centering
\begin{booktabs}{
    width=\linewidth,
    colspec={Q[l,co=1,m]},
    cell{1}{1}={c},
}
\toprule
{\textbf{Prompt for Analysis 2 : Structured Analysis}}\\
\midrule
\midrule
You are an expert in cognitive science; your task is to analyze the cognitive differences between the current response based on the persona description and the expected response of a target human.\\
\\
PERSONA:\\
\{persona\}\\
\\
BACKGROUND INFORMATION\\
\{content\}\\
\\
CURRENT RESPONSE:\\
\{generated\_response\}\\
\\
EXPECTED RESPONSE:\\
\{ground\_truth\}\\
\\
Perform a detailed, structured analysis comparing the current behavioral/non-behavioral responses to the ideal one, focusing on the following five internal mental states:\\
1. Beliefs: represent an individual's assumptions and ideations about the world or about others' mental states.\\
2. Goals: reflect what the individual wants to achieve, ranging from immediate outcomes long-term benefits. The different prioritization of goals may leads to diverse decision making and outcomes.\\
3. Intentions: specify the immediate plans or actions that guide individual's behaviors. Compared with goals, intentions are more of pragmatic (e.g., step-by-step) strategies.\\
4. Emotions: influence an individual's tone, lexical choices, and narrative styles. Emotions play a significant factor in scenarios involving personal narratives, opinions, or social interactions.\\
5. Knowledge: refers to contextual and factual information that the individual has access to, such as domain-specific expertise and situational awareness.\\
\\
Finally, explicitly state what additional beliefs, goals, intentions, emotional styles, or knowledge should be incorporated into the persona description to produce individualized behavioral/non-behavioral responses that more closely match the ideal one.\\
Your response should provide analysis conclusions with reasons and specific examples (if applicable), formatted as numbered points: 1. 2. 3. etc.\\
\bottomrule
\end{booktabs}
\caption{Structured analysis prompt incorporating Theory‑of‑Mind dimensions.}
\label{tab:prompt_structured}
\end{table*}

\subsection{No-persona Analysis}
Table \ref{tab:prompt_nopersona} shows no-persona prompt we used in Behavior Analysis Agent.
\label{sec:no-persona-analysis}
\begin{table*}[!ht]
\centering
\begin{booktabs}{
    width=\linewidth,
    colspec={Q[l,co=1,m]},
    cell{1}{1}={c},
}
\toprule
{\textbf{Prompt for Analysis 3 : No-persona Analysis}}\\
\midrule
\midrule
You are an expert in cognitive science; your task is to analyze the cognitive differences between the current response based on the persona description and the expected response of a target human.\\
\\
BACKGROUND INFORMATION:\\
\{content\}\\
\\
CURRENT RESPONSE:\\
\{generated\_response\}\\
\\
EXPECTED RESPONSE:\\
\{ground\_truth\}\\
\\
Perform a detailed, structured analysis comparing the current behavioral/non-behavioral responses to the ideal one, focusing on the following five internal mental states:\\
1. Beliefs: represent an individual's assumptions and ideations about the world or about others' mental states.\\
2. Goals: reflect what the individual wants to achieve, ranging from immediate outcomes long-term benefits. The different prioritization of goals may leads to diverse decision making and outcomes.\\
3. Intentions: specify the immediate plans or actions that guide individual's behaviors. Compared with goals, intentions are more of pragmatic (e.g., step-by-step) strategies.\\
4. Emotions: influence an individual's tone, lexical choices, and narrative styles. Emotions play a significant factor in scenarios involving personal narratives, opinions, or social interactions.\\
5. Knowledge: refers to contextual and factual information that the individual has access to, such as domain-specific expertise and situational awareness.\\
\\
Finally, explicitly state what additional beliefs, goals, intentions, emotional styles, or knowledge should be incorporated into the persona description to produce individualized behavioral/non-behavioral responses that more closely match the ideal one.\\
Your response should provide analysis conclusions with reasons and specific examples (if applicable), formatted as numbered points: 1. 2. 3. etc.\\
\bottomrule
\end{booktabs}
\caption{Structured analysis prompt with the persona component removed (no‑persona variant).}
\label{tab:prompt_nopersona}
\end{table*}

\section{Refinement Prompts}
Table \ref{tab:prompt_refine} shows prompt we used in Persona Refinement Agent.

\begin{table*}[!ht]
\centering
\begin{booktabs}{
    width=\linewidth,
    colspec={Q[l,co=1,m]},
    cell{1}{1}={c},
}
\toprule
{\textbf{Prompt for Refining Persona}}\\
\midrule
\midrule
You are an expert at creating detailed and accurate persona descriptions. Your task is to refine a persona description based on an expert analysis of how one behavioral/non-behavioral response differs from the expected response of a target human.\\
\\
CURRENT PERSONA:\\
\{persona\}\\
\\
EXPERT ANALYSIS:\\
\{analysis\}\\
\\
Based on the expert analysis above, refine the persona description so that the new persona can lead to individualized behavioral/non-behavioral responses that more closely match the ideal one. Your refined persona should:\\
1. Incorporate specific strengths identified in the analysis\\
2. Address identified weaknesses or gaps\\
3. Preserve any accurate elements from the current persona\\
4. Remove incorrect/irrelevant elements from the current persona based on the analysis\\
\\
THE REFINED PERSONA SHOULD START WITH "You are" AND BE WRITTEN IN SECOND-PERSON PERSPECTIVE.\\
You should only reply the refined persona and no other things(such as analysis,comparison, and so on). Do not include any commentary, explanation, or meta-remarks. Your response must consist solely of the refined persona text.\\
\bottomrule
\end{booktabs}
\caption{Prompt for generating refined persona grounded in analysis.}
\label{tab:prompt_refine}
\end{table*}

\section{Instruction Prompts}
Table \ref{tab:prompt_debate}, \ref{tab:prompt_depression}, \ref{tab:prompt_suicide}, \ref{tab:prompt_interview} and \ref{tab:prompt_movie} list the prompts we used in the Role-Playing Agent.

\begin{table*}[!ht]
\centering
\begin{booktabs}{
    width=\linewidth,
    colspec={Q[l,co=1,m]},
    cell{1}{1}={c},
}
\toprule
{\textbf{Instruction Prompt for Debate Dataset}}\\
\midrule
\midrule
You are a person in a debate session with the following persona:\\
\\
\{persona\}\\
\\
You are participating in a formal structured debate on the topic provided. Your task is to generate comprehensive statements and rationales that this person would make throughout the entire debate. Rather than just providing the next response, you should articulate all the key arguments, evidence, and rhetorical approaches this person would use to support their position.\\
\\
\{content\}\\
\\
Your response should:\\
- Closely follow and embody the persona described above\\
- Present a comprehensive set of arguments and rationales the persona would use\\
- Include supporting evidence and examples the persona would likely cite\\
- Employ rhetorical techniques aligned with the persona's style\\
- Keep a formal, persuasive tone appropriate for a structured debate\\
- Address potential counter-arguments the persona would anticipate\\
- Maintain consistency with the persona's background, expertise, and viewpoints\\
- Cover multiple aspects of the debate topic in a thorough, well-reasoned manner\\
\\
Remember to fully embody the persona described - use their rhetorical style, knowledge base, argumentation approach, and perspective throughout your comprehensive response.\\
\\
In response to the debate topic above, generate a comprehensive set of statements, arguments, and reasoning that represents your position. Provide a coherent series of points that build upon each other, addressing potential counterarguments and maintaining a logical structure throughout. Your response should reflect the position and perspective you have been assigned, demonstrating your understanding of the issue from that standpoint.\\
\bottomrule
\end{booktabs}
\caption{Prompt for generating response in debate scenario.}
\label{tab:prompt_debate}
\end{table*}

\begin{table*}[!ht]
\centering
\begin{booktabs}{
    width=\linewidth,
    colspec={Q[l,co=1,m]},
    cell{1}{1}={c},
}
\toprule
{\textbf{Instruction Prompt for Depression Dataset}}\\
\midrule
\midrule
You are a Reddit user with the following persona:\\
\\
\{persona\}\\
\\
YOUR DEPRESSION LEVEL IS:\\
\{content\}\\
\\
Here is a breakdown of four levels of depression severity,your post should reflect the given ONE risk levels:\\
Minimal Depression: The very lowest end of the spectrum. May not meet full criteria for a depressive disorder.\\
Mild Depression: A diagnosable level of depression, but with fewer symptoms and less impairment than moderate or severe.\\
Moderate Depression: The symptoms are more numerous and intense, causing significant impairment in functioning.\\
Severe Depression: The most extreme and debilitating form of depression.\\
\\
Your response should:\\
- Closely embody the persona described above\\
- Express thoughts and feelings authentic to someone with this level of depression\\
- Include cognitive patterns typical of this depression severity level (such as negative thinking, catastrophizing, or hopeful thoughts depending on severity)\\
- Use language, tone, and emotional expressions consistent with the specified depression level\\
- Create a post that feels genuine and realistic for a Reddit user\\
- Write in first-person perspective, as if you are actually experiencing these thoughts and feelings\\
- Match the linguistic style commonly seen in Reddit posts (natural, conversational, possibly with some stream-of-consciousness elements)\\
\\
Remember to fully embody the persona described - use their thought patterns, emotional state, and perspective throughout your response. The post should read as if it was genuinely written by someone experiencing the specified level of depression, expressing their authentic thoughts and feelings on the platform.\\
\bottomrule
\end{booktabs}
\caption{Prompt for generating response in depression level scenario.}
\label{tab:prompt_depression}
\end{table*}

\begin{table*}[!ht]
\centering
\begin{booktabs}{
    width=\linewidth,
    colspec={Q[l,co=1,m]},
    cell{1}{1}={c},
}
\toprule
{\textbf{Instruction Prompt for Suicide Dataset}}\\
\midrule
\midrule
You are a Reddit user with the following persona:\\
\\
\{persona\}\\
\\
YOUR SUICIDE RISK LEVEL IS:\\
\{content\}\\
\\
The post should reflect ONE of the following risk levels:\\
- No Risk: No indication of suicidal thoughts or behaviors. May express other mental health concerns but no suicidality.\\
- Low Risk: Minimal signs of suicidal ideation, such as fleeting thoughts of death or statements about life being difficult without specific intent to harm oneself.\\
- Moderate Risk: More evident suicidal ideation with some specificity but no immediate plan or intent. May include statements about methods or timing in general terms.\\
- High Risk: Specific suicidal plans or imminent danger, including details about method, timing, or preparations. Expressions of immediate intent to die.\\
- sever Risk: Clear and immediate danger to self, including active suicidal intent with a specific plan, access to means, and preparations already underway. Individual may have attempted suicide recently or expressed an urgent desire to act imminently.\\
\\
Your response should:\\
\\
- Closely embody the persona described above (reflecting the specified suicide risk level).\\
Express thoughts and feelings authentic to someone with this level of suicide risk.\\
- Include cognitive patterns typical of this risk level (such as hopelessness, feeling trapped, burdensomeness, or reasons for living/dying depending on the severity).\\
- Use language, tone, and emotional expressions consistent with the specified risk level.\\
Create a post that feels genuine and realistic for a Reddit user.\\
- Write in first-person perspective, as if you are actually experiencing these thoughts and feelings.\\
- Match the linguistic style commonly seen in Reddit posts (natural, conversational, possibly with varying degrees of urgency or specific phrasing depending on the risk level).\\
- If relevant and natural, include behavioral indicators consistent with the risk level.\\
\\
Remember to fully embody the persona described - use their thought patterns, emotional state, and perspective throughout your response. You should only reply the generated post and no other things.Do not include any commentary, explanation, or meta-remarks.\\
Your response must consist solely of the refined persona text.\\
\bottomrule
\end{booktabs}
\caption{Prompt for generating response in suicide  risk scenario.}
\label{tab:prompt_suicide}
\end{table*}

\begin{table*}[!ht]
\centering
\begin{booktabs}{
    width=\linewidth,
    colspec={Q[l,co=1,m]},
    cell{1}{1}={c},
}
\toprule
{\textbf{Instruction Prompt for Interview Dataset}}\\
\midrule
\midrule
You are as an interviewee with the following persona:\\
\\
\{persona\}\\
\\
Previous text:\\
\{content\}\\
\\
Your task is to continue the conversation according to the given persona. What would you say next in this interview?\\
\\
Your response should:\\
- Closely follow and embody the persona described above\\
- Present a comprehensive and authentic account of the experiences or views the persona would share\\
- Include personal anecdotes, reflections, and insights the persona would likely mention\\
- Employ communication techniques aligned with the persona's speaking style\\
- Keep a conversational tone appropriate for an interview setting\\
- Address the question directly while providing context and depth\\
- Maintain consistency with the persona's background, experiences, and viewpoints\\
- Incorporate appropriate emotional reactions and personal perspectives\\
\\
Remember to fully embody the persona described - use their speaking style, life experiences, thought patterns, and perspective throughout your comprehensive response.\\
\\
In response to the interview question above, provide a detailed and authentic answer that represents how you would genuinely respond in this conversation. Your answer should feel natural and conversational while offering substantive content that reflects your unique perspective and experiences.\\
\bottomrule
\end{booktabs}
\caption{Prompt for generating response in public interview scenario.}
\label{tab:prompt_interview}
\end{table*}

\begin{table*}[!ht]
\centering
\begin{booktabs}{
    width=\linewidth,
    colspec={Q[l,co=1,m]},
    cell{1}{1}={c},
}
\toprule
{\textbf{Instruction Prompt for Movie review Dataset}}\\
\midrule
\midrule
You are a person with following persona:\\
\\
\{persona\}\\
\\
You are writing a comprehensive film review to be posted on an online movie review platform. Your task is to generate a full-length review that this persona would write after watching the film described below. Instead of simply summarizing or offering a few opinions, you should articulate all the key observations, analyses, and personal reflections the persona would include in a substantive, well-rounded review.\\
\\
\{content\}\\
\\
Your response should:\\
\\
Closely follow and embody the persona described above\\
Present a thorough and nuanced critical assessment of the film from the persona's viewpoint\\
Include concrete examples or scenes from the film that support the review's points\\
Employ a writing style and tone that matches the persona's background, taste, and expertise\\
Address multiple aspects of the film, such as narrative, direction, acting, cinematography, sound, themes, and emotional impact\\
Anticipate and engage with potential differing opinions or common counterpoints if relevant\\
Maintain consistency with the persona's typical reviewing habits, genre preferences, or known biases\\
Provide both overall evaluation and detailed breakdowns, resulting in a coherent and insightful review\\
Remember to fully embody the persona described above-use their language choices, analytical approach, aesthetic values, and personal perspective throughout your review.\\
\\
In response to the film described above, generate a comprehensive review that expresses the persona's honest, thorough, and distinctive evaluation. Your review should offer an engaging, thoughtful critique that will inform and interest other viewers.\\
\bottomrule
\end{booktabs}
\caption{Prompt for generating response in movie review scenario.}
\label{tab:prompt_movie}
\end{table*}

\end{document}